\documentclass[10pt,twocolumn,letterpaper]{article}

\usepackage{cvpr}
\usepackage{times}
\usepackage{epsfig}
\usepackage{graphicx}
\usepackage{amsmath}
\usepackage{amssymb}
\usepackage{booktabs}
\usepackage{balance}
\usepackage{lscape}
\usepackage{multirow}

\usepackage[flushleft]{threeparttable}
\usepackage{arydshln}
\usepackage{bm}
\usepackage{caption}
\usepackage{subcaption}
\usepackage{afterpage}

\usepackage[pagebackref=true,breaklinks=true,letterpaper=true,colorlinks,bookmarks=false]{hyperref}

\cvprfinalcopy 

\def\cvprPaperID{7812} 

\ifcvprfinal\fi
\begin{document}

\title{Assessing Image Quality Issues for Real-World Problems}

\author{\hspace{-0.1in} Tai-Yin Chiu, Yinan Zhao, Danna Gurari \\ \noindent {\hspace{-0.2in} \small University of Texas at Austin} }

\maketitle

\begin{abstract}
We introduce a new large-scale dataset that links the assessment of image quality issues to two practical vision tasks: image captioning and visual question answering.  First, we identify for 39,181 images taken by people who are blind whether each is sufficient quality to recognize the content as well as what quality flaws are observed from six options.  These labels serve as a critical foundation for us to make the following contributions: (1) a new problem and algorithms for deciding whether an image is insufficient quality to recognize the content and so not captionable, (2) a new problem and algorithms for deciding which of six quality flaws an image contains, (3) a new problem and algorithms for deciding whether a visual question is unanswerable due to unrecognizable content versus the content of interest being missing from the field of view, and (4) a novel application of more efficiently creating a large-scale image captioning dataset by automatically deciding whether an image is insufficient quality and so should not be captioned.  We publicly-share our datasets and code to facilitate future extensions of this work: \texttt{https://vizwiz.org}. 
\end{abstract}


\section{Introduction}
Low-quality images are an inevitable, intermittent reality for many real-world, computer vision applications.  At one extreme, they can be life threatening, such as when they impede the ability of autonomous vehicles \cite{zhu2016traffic} and traffic controllers \cite{lou2019veri} to safely navigate environments.  In other cases, they can serve as irritants when they convey a negative impression to the viewing audiences, such as on social media or dating websites.  

Despite that low-quality images often emerge in practical settings, there has largely been a disconnect between research aimed at recognizing quality issues and research aimed at performing downstream vision tasks.  For researchers focused on uncovering what quality issues are observed in an image, their progress largely has grown from artificially-constructed settings where they train and evaluate algorithms on publicly-available datasets that were constructed by distorting high quality images to simulate quality issues (e.g., using JPEG compression or Gaussian blur)~\cite{sheikh2006statistical, wang2004image, ghadiyaram2015massive, jayaraman2012objective, ponomarenko2009tid2008, ponomarenko2015image, larson2010most, ma2016waterloo}.  Yet, these contrived environments typically lack sufficient sophistication to capture the plethora of factors that contribute to quality issues in natural settings (e.g., camera hardware, lighting, camera shake, scene obstructions).  Moreover, the quality issues are detangled from whether they relate to the ability to complete specific vision tasks.  As for researchers focusing on specific tasks, much of their progress has developed from environments that lack low-quality images.  That is because the creators of popular publicly-available datasets that support the development of such algorithms typically included a step to filter out any candidate images that are deemed insufficient quality for the final dataset~\cite{fei2004learning, griffin2007caltech, krizhevsky2009learning, deng2009imagenet, xiao2010sun, lin2014microsoft, zhou2017places}.  Consequently, such datasets lack data that would enable training algorithms to identify when images are of insufficient quality to complete a given task.

\begin{figure*}[t!]
    \centering
    \includegraphics[width=\linewidth]{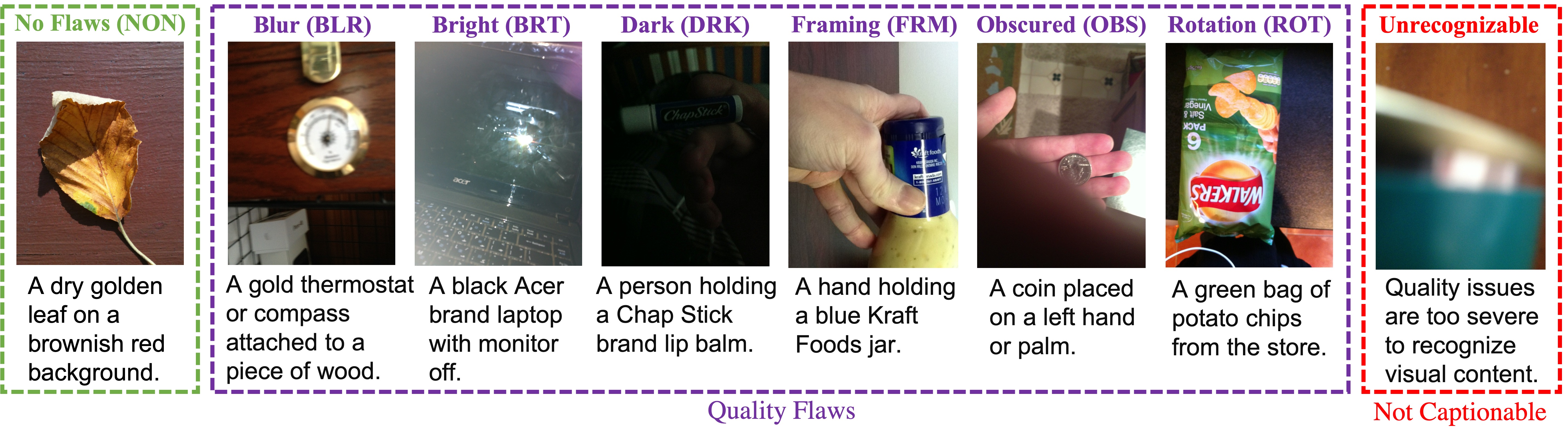}
    \vspace{-1.5em}
    \caption{We introduce a new image quality assessment dataset which we call VizWiz-QualityIssues.  Shown are examples of the taxonomy of labels, which ranges from no quality issues to six quality flaws to unrecognizable/uncaptionable images.  Images can manifest different combinations of the above labels, for instance the unrecognizable image is also labeled as suffering from image blur and poor framing.}
    \label{fig:exampleImages}
\end{figure*}

Motivated by the aim to tie the assessment of image quality to practical vision tasks, we introduce a new image quality assessment (IQA) dataset that emerges from a real use case.  Specifically, our dataset is built around 39,181 images that were taken by people who are blind who were authentically trying to learn about images they took using the VizWiz mobile phone application~\cite{bigham2010vizwiz}.  Of these images, 17\% were submitted to collect image captions from remote humans.  The remaining 83\% were submitted with a question to collect answers to their visual questions.  As discussed in prior work~\cite{brady2013visual,gurari2018vizwiz}, users submitted these images and visual questions (i.e., images with questions) to overcome real visual challenges that they faced in their daily lives.  They typically waited nearly two minutes to receive a response from the remote humans~\cite{bigham2010vizwiz}.  For each image, we asked crowdworkers to either supply a caption describing it or clarify that the quality issues are too severe for them to be able to create a caption.  We call this task the \emph{unrecognizability classification task}.  We also ask crowdworkers to label each image with quality flaws that are more traditionally discussed in the literature~\cite{brady2013visual,ghadiyaram2015massive}: blur, overexposure (bright), underexposure (dark), improper framing, obstructions, and rotated views.  We call this task the \emph{quality flaws classification task}.  Examples of resulting labeled images in our dataset are shown in Figure~\ref{fig:exampleImages}.  Altogether, we call this dataset VizWiz-QualityIssues.

We then demonstrate the value of this new dataset for several new purposes.  First, we introduce a novel problem and algorithms for predicting whether an image is sufficient quality to be captioned (Section~\ref{sec_classifyingQualityIssues}).  This can be of immediate use to blind photographers, who otherwise must wait nearly two minutes to learn their image is unsuitable quality for image captioning.  We next conduct experiments to demonstrate an additional benefit of this prediction system for creating large-scale image captioning datasets with less wasted human effort (Section~\ref{sec_budgetAllocation}).  Finally, we introduce a novel problem and algorithms that inform a user who submits a novel visual question whether it can be answered, cannot be answered because the image content is unrecognizable, or cannot be answered because the image content is missing from the image (Section~\ref{sec_vqa}).  This too can be of immediate benefit to blind photographers by enabling them to both fail fast and gain valuable insight into how to update the visual question to make it become answerable. 

More generally, our work underscores the importance of defining quality within the context of specific tasks.  We expect our work can generalize to related vision tasks such as object recognition, scene classification, and video analysis. 
\section{Related Work}
\label{sec:related_work}

\paragraph{Image Quality Datasets.} 
A number of image quality datasets exist to support the development of image quality assessment (IQA) algorithms, including LIVE \cite{sheikh2006statistical, wang2004image}, LIVE MD \cite{jayaraman2012objective}, TID2008 \cite{ponomarenko2009tid2008}, TID2013 \cite{ponomarenko2015image}, CSIQ \cite{larson2010most}, Waterloo Exploration~\cite{ma2016waterloo}, and ESPL-LIVE\cite{kundu2017large}. A commonality across most such datasets is that they originate from high quality images that were artificially distorted to introduce image quality issues. For example, LIVE~\cite{ghadiyaram2015massive} consists of 779 distorted images, which are derived by applying five different types of distortions at numerous distortion levels to 29 high-quality images.  Yet, image quality issues that arise in real-world settings exhibit distinct appearances than those that are found by simulating distortions to high-quality images.  Accordingly, our work complements recent efforts to create large-scale datasets that flag quality issues in natural images~\cite{ghadiyaram2015massive}.  However, our dataset is considerably larger, offering approximately a 19-fold increase in the number of naturally distorted images; i.e., 20,244 in our dataset versus 1,162 images for \cite{ghadiyaram2015massive}.  In addition, while \cite{ghadiyaram2015massive} assigns a single quality score to each image to capture any of a wide array of image quality issues, our work instead focuses on recognizing the presence of each distinct quality issue and assessing the impact of the quality issues on the real application needs of real users.

\vspace{-0.7em}
\paragraph{Image Quality Assessment.} 
Our work also relates to the literature that introduces methods for assessing the quality of images.  One body of work assumes that developers have access to a high-quality version of each novel image, whether partially or completely.  For example, distorted images are evaluated against original, intact images for full-reference IQA algorithms \cite{wang2004image, wang2003multiscale, zhang2011fsim, sheikh2006statistical, larson2010most, bosse2017deep, reisenhofer2018haar} and distorted images are evaluated against partial information about the original, intact images for reduced-reference IQA algorithms \cite{wang2005reduced, li2009reduced, wang2011reduced, soundararajan2011rred, rehman2012reduced, ma2011reduced, wu2013reduced}.  Since our natural setting inherently limits us from having access to original, intact images, our work instead aligns with the second body of work which is built around the assumption that no original, reference image is available; i.e., no-reference IQA (NR-IQA).  NR-IQA algorithms instead predict a quality score for each novel image~\cite{mittal2012no, kang2014convolutional, wang2011reduced, ye2012unsupervised, ye2012no, liu2014no, suresh2009no, bosse2017deep, talebi2018nima}.  While many algorithms have been introduced for this purpose, our analysis of five popular NR-IQA models (i.e., BRISQUE \cite{mittal2012no}, NIQE \cite{mittal2012making}, CNN-NRIQA \cite{kang2014convolutional}, DNN-NRIQA \cite{bosse2017deep}, and NIMA \cite{talebi2018nima}) demonstrates that they are inadequate for our novel task of assessing which images are unrecognizable and so cannot be captioned (discussed in Section~\ref{sec_classifyingQualityIssues}).  Accordingly, we introduce new algorithms for this purpose, and demonstrate their advantage.

\vspace{-0.7em}
\paragraph{Efficient Creation of Large-Scale Vision Datasets.} 
Progress in the vision community has largely been measured and accelerated by the creation of large-scale vision datasets over the past 20 years.  Typically, researchers have scraped images for such datasets from online image search databases~\cite{fei2004learning, griffin2007caltech, krizhevsky2009learning, deng2009imagenet, xiao2010sun, lin2014microsoft, zhou2017places}.  In doing so, they typically curate a large collection of high-quality images, since such images first passed uploaders' assessment that they are of sufficient quality to be shared publicly.  In contrast, when employing images captured ``in the wild," it can be a costly, time-consuming process to identify and remove images with unrecognizable content.  Accordingly, we quantify the cost of this problem, introduce a novel problem and algorithms for deciphering when image content would be unrecognizable to a human and so should be discarded, and demonstrate the benefit of such solutions for more efficiently creating a large-scale image captioning dataset.

\vspace{-0.7em}
\paragraph{Assistive Technology for Blind Photographers.}
Our work relates to the literature about technology for assisting people who are blind to take high-quality pictures~\cite{taptapseeapp,bigham2010vizwiz,jayant2011supporting,vazquez2014assisted,zhong2013real}.  Already, existing solutions can assist photographers in improving the image focus~\cite{taptapseeapp}, lighting~\cite{bigham2010vizwiz}, and composition~\cite{jayant2011supporting,vazquez2014assisted,zhong2013real}.  Additionally, algorithms can inform photographers whether their questions about their images can be answered~\cite{gurari2018vizwiz} and why crowds struggle to provide answers~\cite{bhattacharya2019does,gurari2017crowdverge}.  Complementing prior work, we introduce a suite of new AI problems and solutions for offering more fine-grained guidance when alerting blind photographers about what image quality issue(s) are observed.  Specifically, we introduce novel problems of (1) recognizing whether image content can be recognized (and so captioned) and (2) deciphering when a question about an image can be answered, cannot be answered because the image content is unrecognizable, or cannot be answered because the content of interest is missing from the image.  
\section{VizWiz-QualityIssues}
We now describe our creation of a large-scale, human-labeled dataset to support the development of algorithms that can assess the quality of images.  We focus on a real use case that is prone to image quality issues.  Specifically, we build off of 39,181 publicly-available images~\cite{gurari2019vizwiz,gurari2018vizwiz} that originate from blind photographers who each submitted an image with, optionally, a question to the VizWiz mobile phone application~\cite{bigham2010vizwiz} in order to receive descriptions of the image from remote humans.  Since blind photographers are unable to verify the quality of the images they take, the dataset exemplifies the large diversity of quality issues that occur naturally in practice.  We describe below how we create and analyze our new dataset.
 
\subsection{Creation of the Dataset}
We scoped our dataset around quality issues that impede people who are blind in their daily lives.  Specifically, a clear, resounding message is that people who are blind need assistance in taking images that are sufficiently high-quality that sighted people are able to either describe them or answer questions about them~\cite{bigham2010vizwiz,brady2013visual}.

\vspace{-0.7em}
\paragraph{Quality Issues Taxonomy.}
One quality issue label we assess is whether \emph{image content is sufficiently recognizable} for sighted people to caption the images.  We also label numerous \emph{quality flaws} to situate our work in relation to other papers that similarly focus on assessing image quality issues~\cite{brady2013visual,ghadiyaram2015massive}.  Specifically, we include the following categories: blur (is the image blurry?), bright (is the image too bright?), dark (is the image too dark?), obstruction (is the scene obscured by the photographer's finger over the lens, or another unintended object?), framing (are parts of necessary items missing from the image?), rotation (does the image need to be rotated for proper viewing?), other, and no issues (there are no quality issues in the image). 

\vspace{-0.7em}
\paragraph{Image Labeling Task.}
To efficiently label all images, we designed our task to run on the crowdsourcing platform Amazon Mechanical Turk.  The task interface showed an image on the left half and the instructions with user-entry fields on the right half.  First, the crowdworker was instructed to either describe the image in one sentence or click a button to flag the image as being insufficient quality to recognize the content (and so not captionable).  When the button was clicked, the image description was automatically populated with the following text: ``Quality issues are too severe to recognize the visual content."  Next, the crowdworker was instructed to select all image quality flaws from a pre-defined list that are observed.  Shown were the six reasons identified above, as well as Other (OTH) linked to a free-entry text-box so other flaws could be described and None (NON) so crowd workers could specify the image had no quality flaws.  The interface enabled workers to adjust their view of the image, using the toolbar to zoom in, zoom out, pan around, or rotate the image if needed. To encourage higher quality results, the interface prevented a user from completing the task until a complete sentence was provided and at least one option from the ``image quality flaw" options was chosen.  A screen shot of the user interface is shown in the Supplementary Materials.  
 
\vspace{-0.7em}
\paragraph{Crowdsourcing Labels.}
To support the collection of high quality labels, we only accepted crowdworkers who previously had completed over 500 HITs with at least a 95\% acceptance rate.  Also, we collected redundant results.  Specifically, we recruited five crowdworkers to label each image.  We deemed a label as valid only if at least two crowdworkers chose that label.

\subsection{Characterization of the Dataset}
\paragraph{Prevalence of Quality Issues.}
We first examine the frequency at which images taken by people who are blind suffer from the various quality issues to identify the (un)common reasons.  To do so, we tally how often unrecognizable images and each quality-flaw arise.  

Roughly half of the images suffer from image quality flaws (i.e., 1-$P$(NON)=$51.6\%$).  We observe that the most common reasons are image blur (i.e., $41.0\%$) and inadequate framing (i.e., $55.6\%$).  In contrast, only a small portion of the images are labeled as too bright (i,e., $5.3\%$), too dark ($5.6\%$), having objects obscuring the scene ($3.6\%$), needing to be rotated for successful viewing ($17.5\%$), or other reasons ($0.8\%$).  The statistics reveal the most promising directions for how to improve assistive photography tools to improve blind users' experiences.  Specifically, the main functions should be focused on camera shake detection and object detection to mitigate the possibility of taking images with blur or framing flaws. 

We also observe that the image quality issues are so severe that image content is deemed unrecognizable for 14.8\% of the images.  In absolute terms, this means that \$3,829 and 379 hours of human annotation were wasted on employing crowdworkers to caption images that contained unrecognizable content.\footnote{Crowdworkers were paid \$0.132 for each image and spent an average of 47 seconds captioning each image.}  In other words, great savings can be achieved by automatically filtering such uncaptionable images such that they are not sent to crowdworkers.  We explore this idea further in Section~\ref{sec_budgetAllocation}.

\begin{figure}[b!]
    \centering
    \vspace{-0.2em}
    \includegraphics[width=\linewidth]{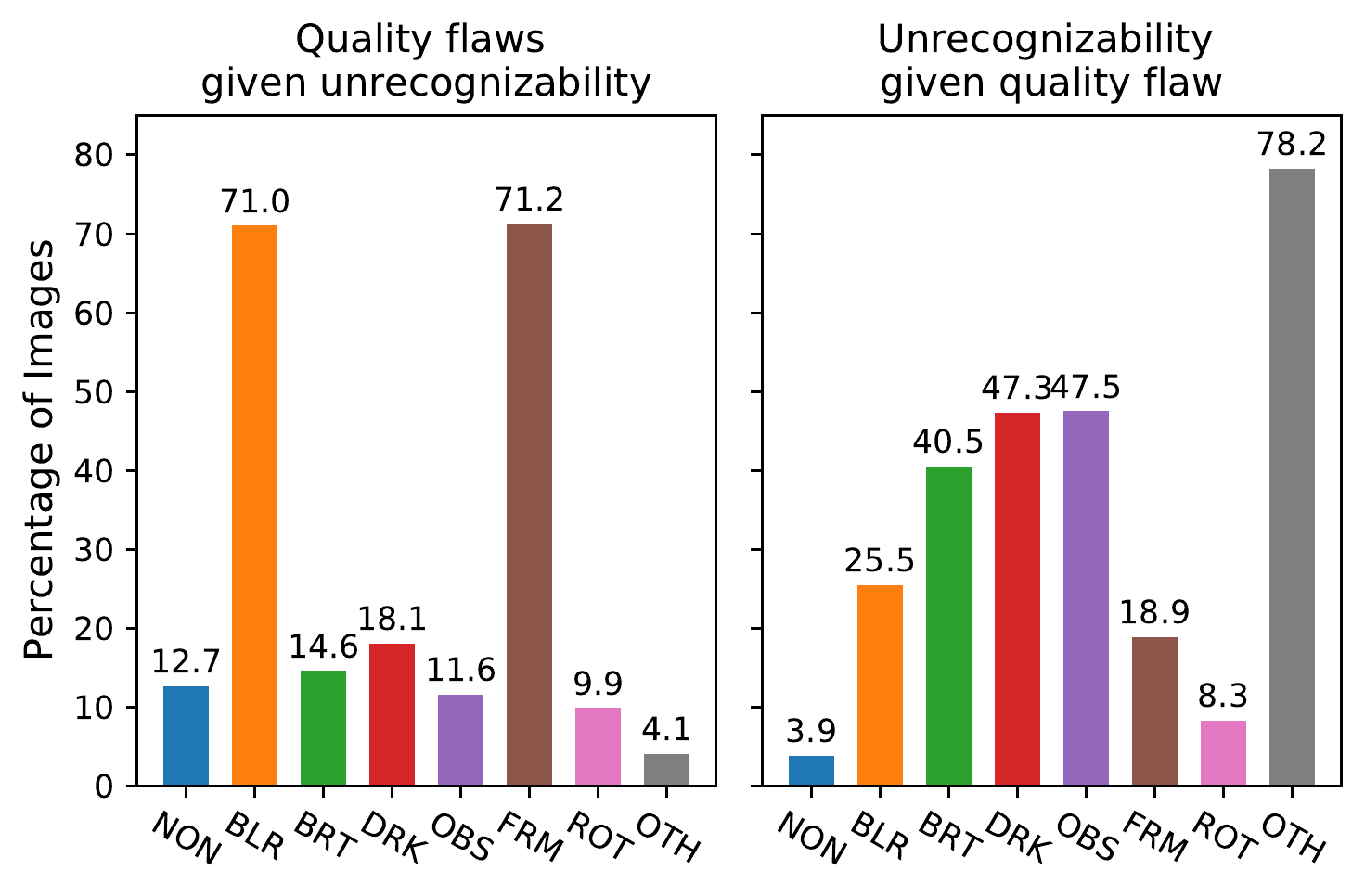}\vspace{-2ex}
    \caption{Left: Percentage of images with quality flaws given unrecognizability. Right: Percentage of unrecognizable images given quality flaws.}
    \label{fig:unrecognizable}
\end{figure}

\vspace{-0.7em}
\paragraph{Likelihood Image Has Unrecognizable Content Given its Quality-Flaw.}
We next examine the probability that an image's content is unrecognizable conditioned on each of the reasons for quality flaws.  Results are shown in Figure~\ref{fig:unrecognizable}. 

Almost all reasons led to percentages that are larger than the overall percentage of unrecognizable images, which is $14.8\%$ of all images.  This demonstrates what we intuitively suspected, which is that images with quality flaws are more likely to have unrecognizable content.  We observe that this trend is the strongest for images that suffer from obstructions (OBS) and inadequate lighting (BRT and DRK), with percentages just over 40\%.

Interestingly, two categories have percentages that are smaller than the overall percentage of unrecognizable images, at $14.8\%$ of all images.  First, images that are flagged as needing to be rotated for proper viewing (ROT) have only $8.3\%$ deemed unrecognizable.  In retrospect, this seems understandable, as the content of images with a rotation flaw could still be recognized if viewers tilt their heads (or apply visual display tools to rotate the images).   Second, images labeled with no flaws (NON) have only $3.9\%$ deemed unrecognizable.  This tiny amount aligns with the concept that ``unrecognizable" and ``no flaws" are two conflicting ideas.  Still, the fact the percentage is not 0\% highlights that humans can offer different perspectives.  Put differently, the image quality assessment task can be subjective.

\begin{figure}[b!]
     \centering
     \includegraphics[width=0.48\textwidth]{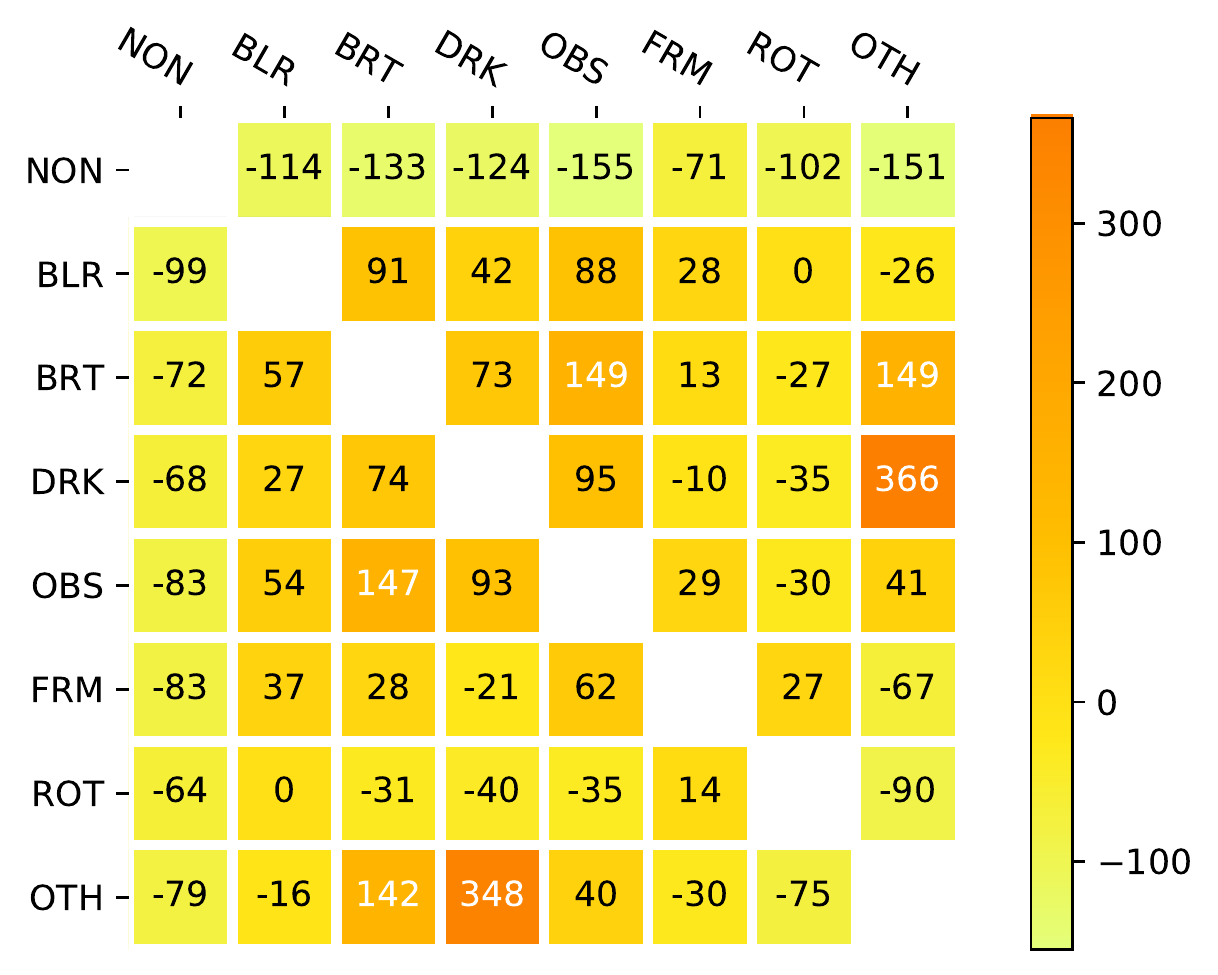}
     \vspace{-1.8em}
     \caption{Interrelation of quality flaws. Values are scaled, with each multiplied by 100. The grid at the $i$-th row and the $j$-th column shows the value of $I(\textnormal{flaw }i, \textnormal{flaw }j)$. The diagonal is suppressed for clarity.}
     \label{fig:quality_issue_coocc}
\end{figure}

\begin{figure*}[t!]
     \centering
     \begin{subfigure}[b]{0.16\textwidth}
         \centering
         \includegraphics[width=\textwidth]{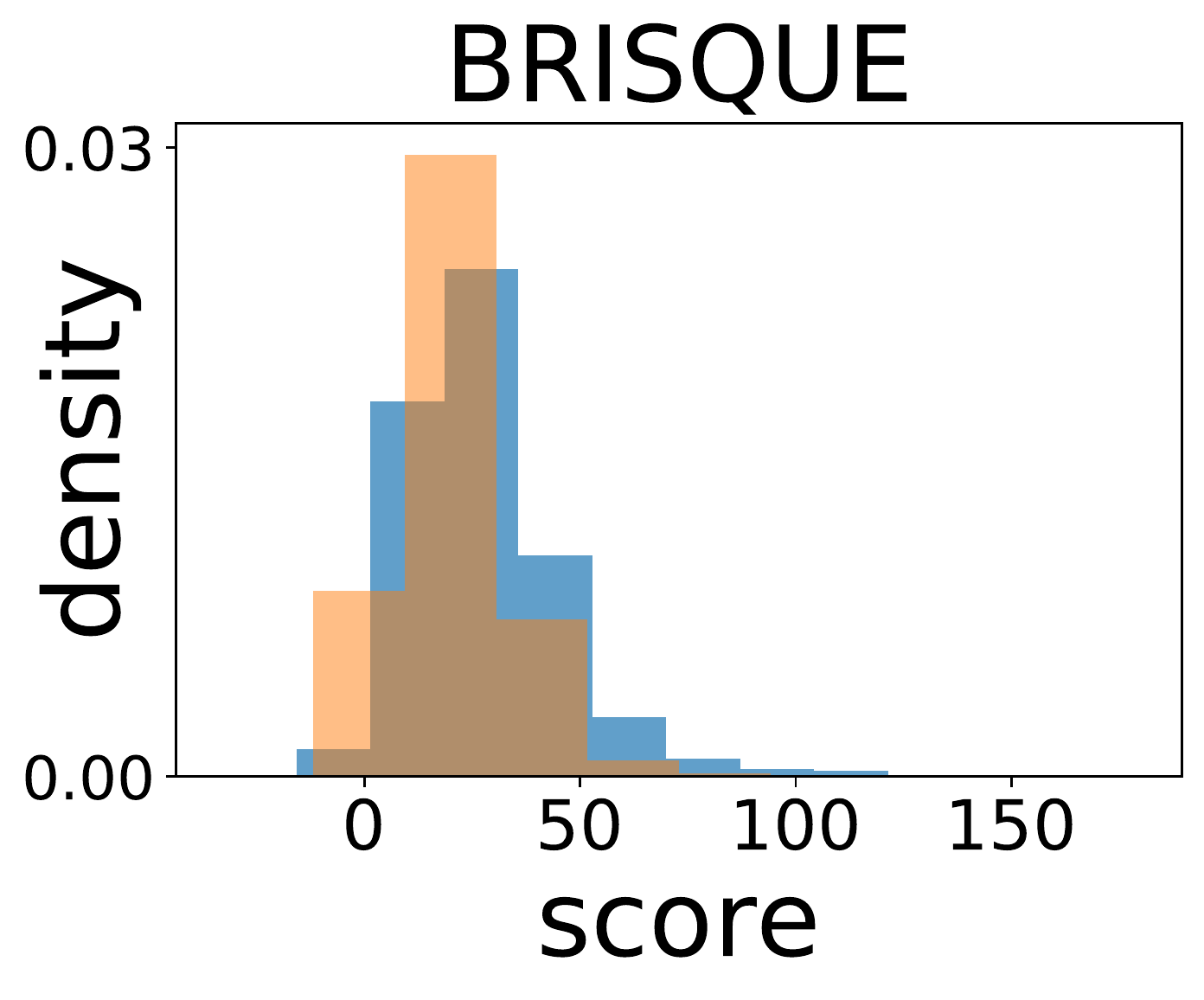}
         \label{fig:iqa_brisque}
     \end{subfigure}
     \hfill
     \begin{subfigure}[b]{0.16\textwidth}
         \centering
         \includegraphics[width=\textwidth]{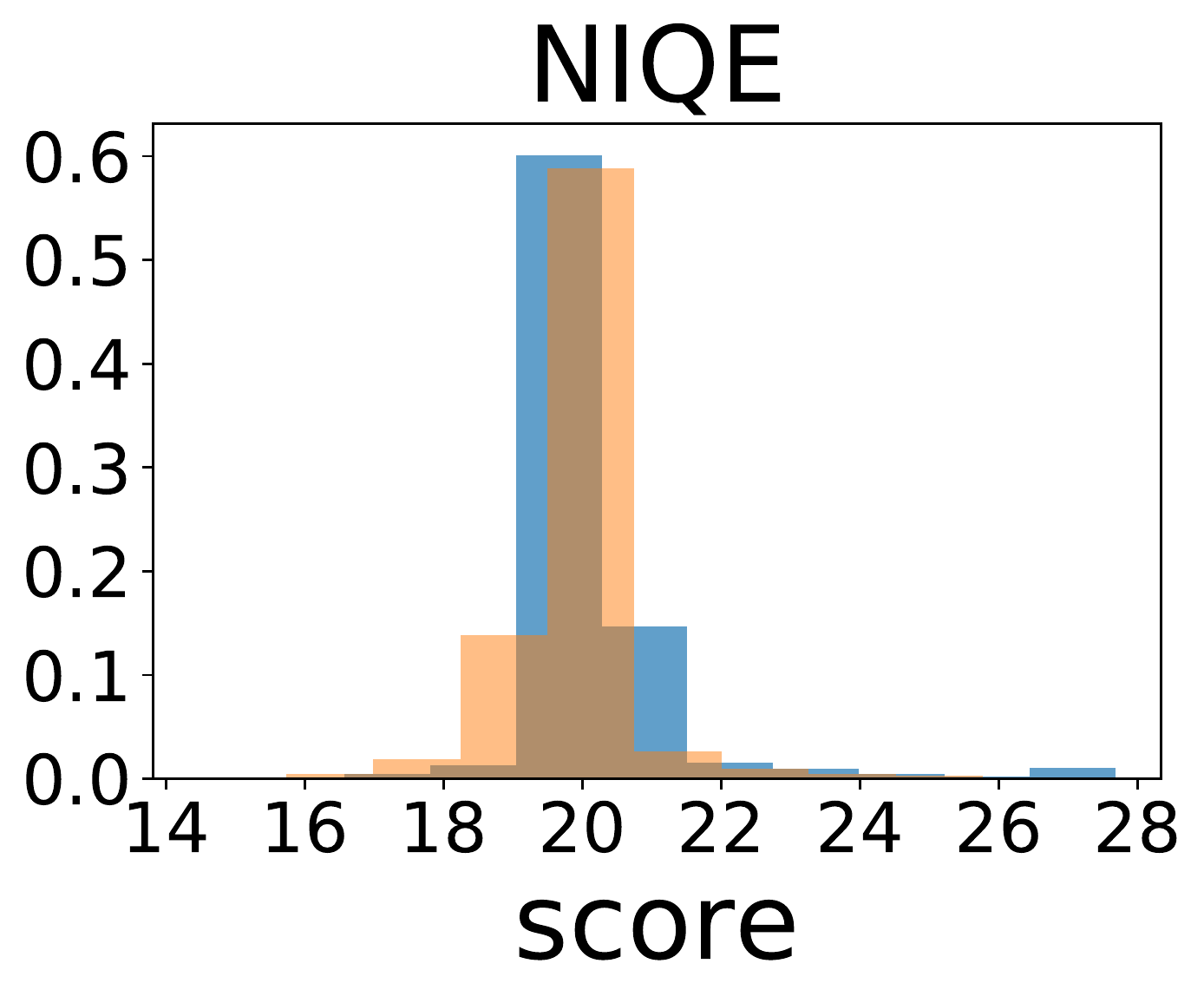}
         \label{fig:iqa_niqe}
     \end{subfigure}
     \hfill
     \begin{subfigure}[b]{0.16\textwidth}
         \centering
         \includegraphics[width=\textwidth]{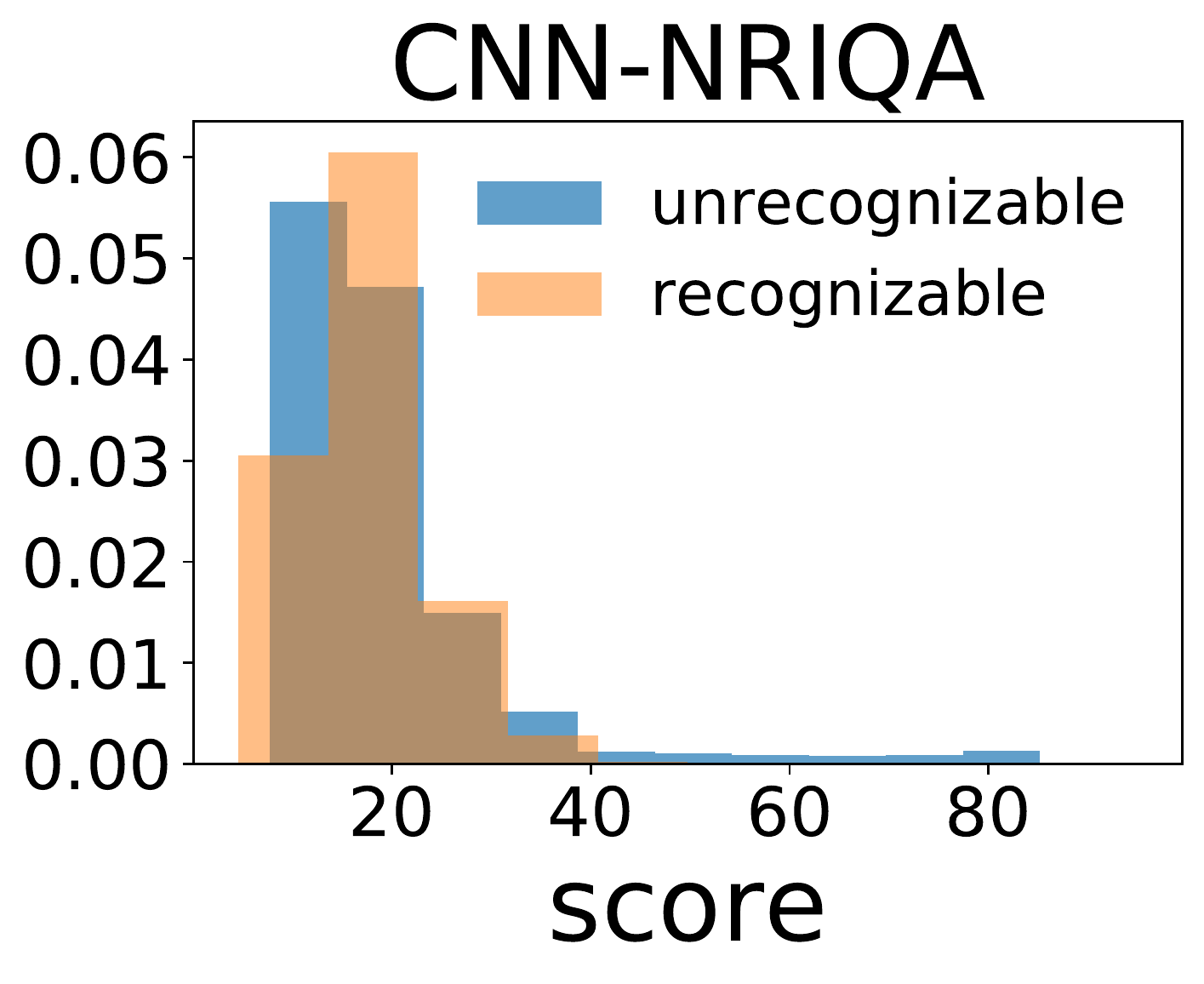}
         \label{fig:iqa_cnn_nriqa}
     \end{subfigure}
     \hfill
     \begin{subfigure}[b]{0.16\textwidth}
         \centering
         \includegraphics[width=\textwidth]{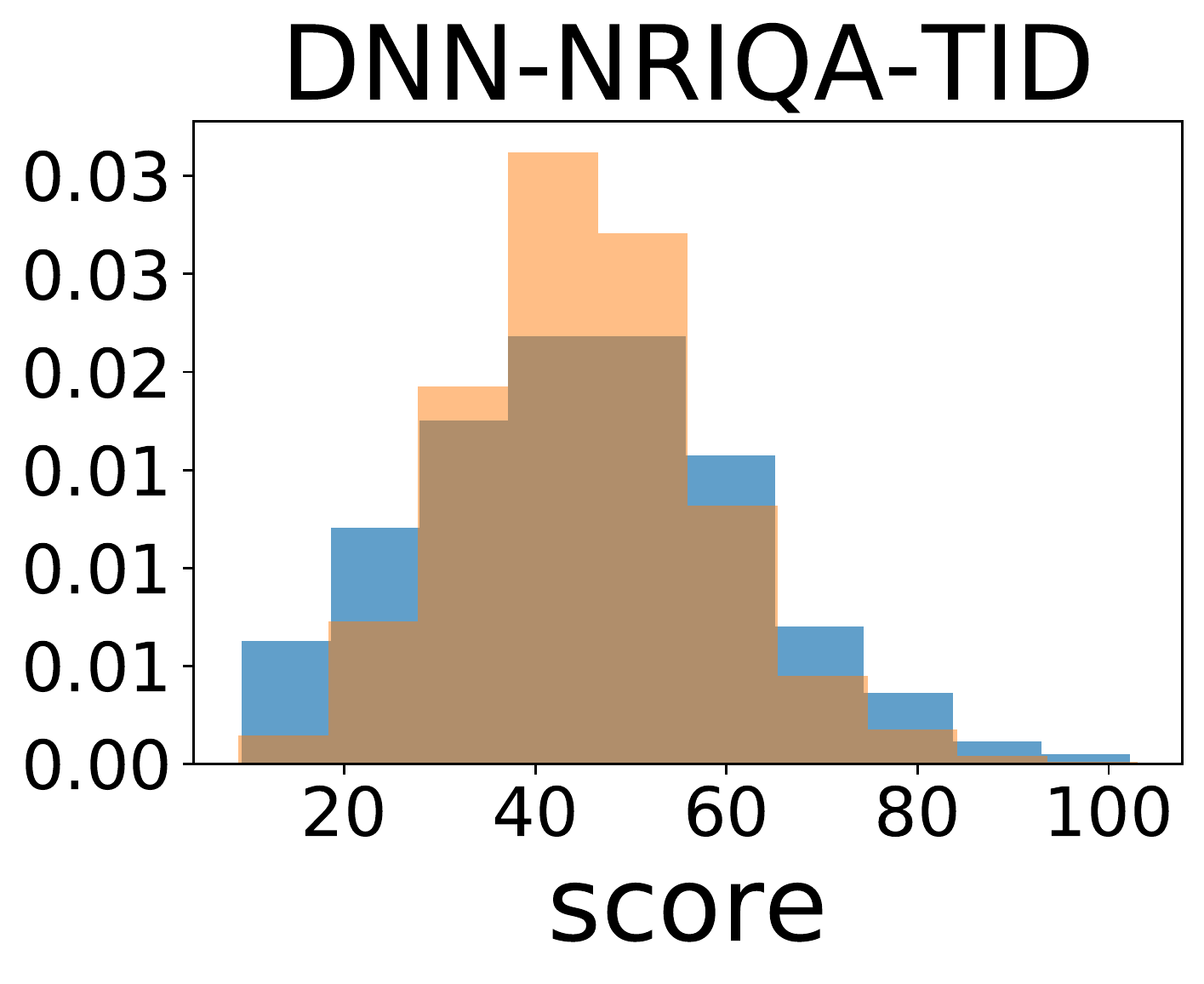}
         \label{fig:iqa_dnn_nriqa_tid}
     \end{subfigure}
     \hfill
     \begin{subfigure}[b]{0.16\textwidth}
         \centering
         \includegraphics[width=\textwidth]{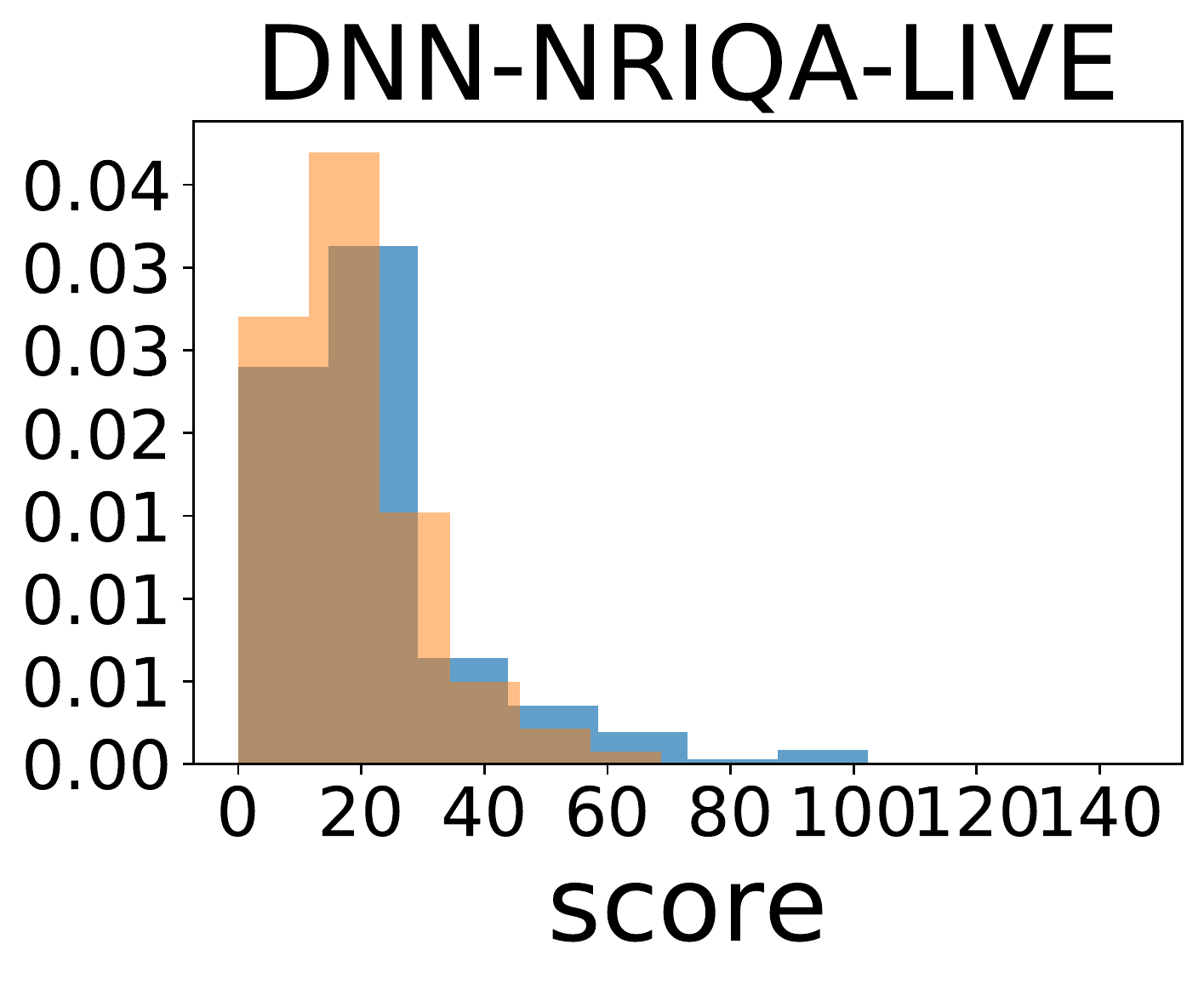}
         \label{fig:iqa_dnn_nriqa_live}
     \end{subfigure}
     \hfill
     \begin{subfigure}[b]{0.16\textwidth}
         \centering
         \includegraphics[width=\textwidth]{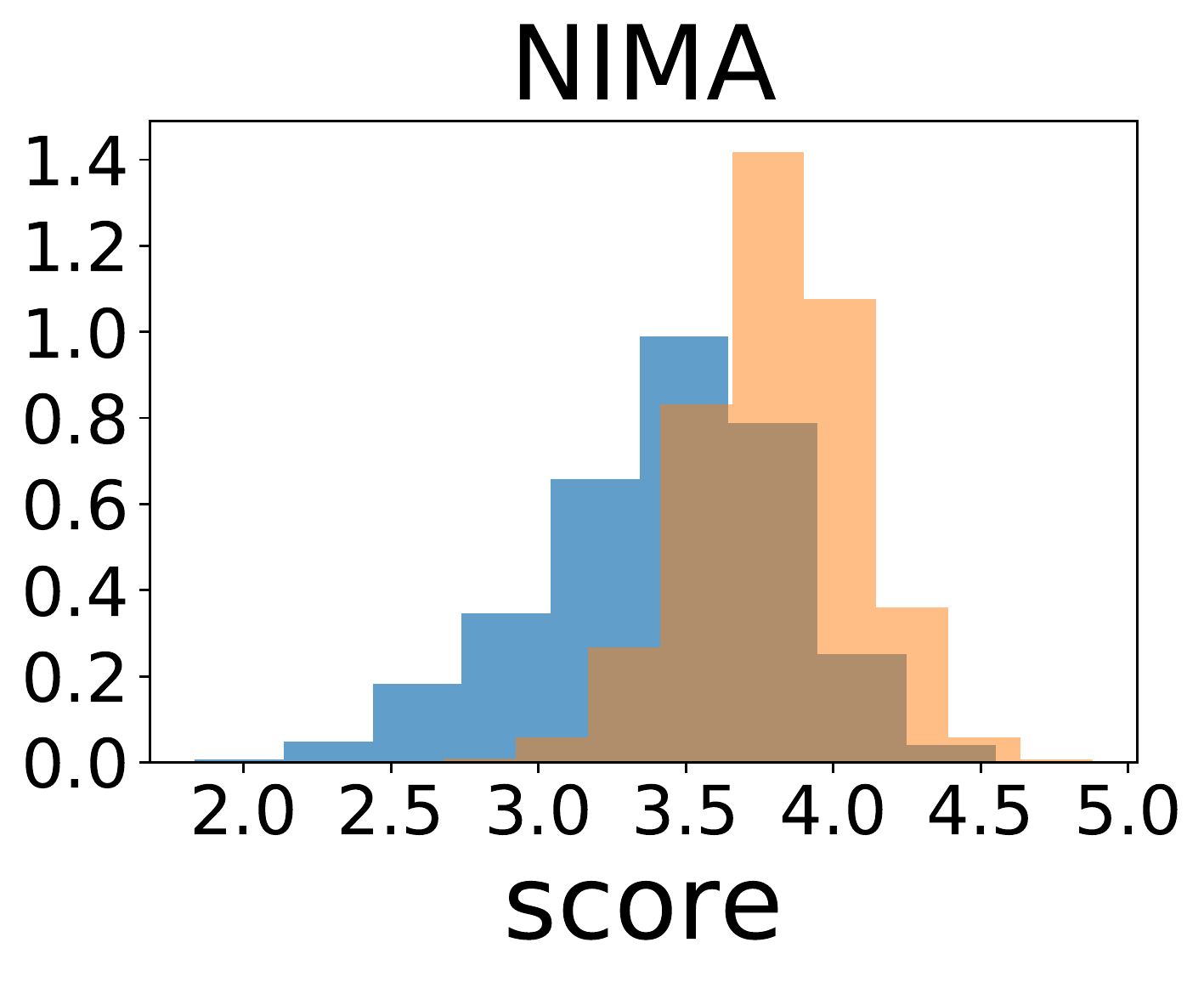}
         \label{fig:iqa_nima}
     \end{subfigure}\vspace*{-6mm}
     \caption{Distributions of image quality scores predicted by conventional NR-IQA systems \cite{mittal2012no, mittal2012making, kang2014convolutional, bosse2017deep, talebi2018nima} in our new VizWiz-QualityIssues dataset.  The heavy overlap of the distributions of scores for recognizable and unrecognizable images reveals that none of the methods are able to distinguish recognizable images from unrecognizable images.}
     \label{fig:NRIQA}
\end{figure*}

\vspace{-0.7em}
\paragraph{Likelihood Image Has Each Quality-Flaw Given its Content is Unrecognizable.}
We next examine the probability that an image manifests each quality flaw given that its content is unrecognizable.  Results are shown in Figure~\ref{fig:unrecognizable}.  Overall, our findings parallel those identified in the ``Prevalence of Quality Issues" paragraph.  For example, we again observe the most common reasons are blurry images ($71.0\%$) and improper framing (71.2\%).  Similarly, unrecognizable images are found to be associated less frequently with the other quality flaws.

\vspace{-0.7em}
\paragraph{Relationship Between Quality Flaws in Images.}
Finally, we quantify the relationship between all possible pairs of quality flaws.  In doing so, we were motivated to provide a measure that offers insight into causality and co-occurrence when comparing any pair of quality flaws, while avoiding measuring joint probabilities.  To meet this aim, we introduce a new measure which we call interrelation index $I(A,B)$, which is defined as follows:
\begin{equation}
    I(A,B)=\frac{P(B|A)}{P(B)} - \frac{P(B|\bar{A})}{P(B)}.
    \label{eq:interrelation}
\end{equation}
 
 \noindent
 More details about this measure and the motivation for it are provided in the Supplementary Materials.
 Briefly, larger positive $I(A,B)$ values indicate that $A$ and $B$ tend to co-occur with $A$ causing $B$ to happen more often.  Results are shown in Figure~\ref{fig:quality_issue_coocc}.  
 
We observe that almost all quality flaws tend to occur with one another, as shown with the positive values of $I$.  At first, we were surprised to observe that there is a relationship between BRT and DRK (i.e., $I(\textnormal{BRT},\textnormal{DRK})=73$ is greater than zero), since these flaws are seemingly incompatible concepts.  However, from visual inspection of the data, we found some images indeed suffered from both lighting flaws.  We exemplify this and other quality flaw correlations in the Supplementary Materials.  From our findings, we also observe that ``no flaws" does not co-occur with other quality flaws; i.e., the values in the grid are all negative for the row and column for NON.  This finding aligns with our intuition that an image labeled with NON is less likely to have a quality flaw at the same time. 
\section{Classifying Unrecognizable Images}
\label{sec_classifyingQualityIssues}
A widespread assumption when captioning images is that the image quality is good enough to recognize the image content.  Yet, people who are blind cannot verify the quality of the images they take and it is known their images can be very poor in quality~\cite{bigham2010vizwiz,brady2013visual,gurari2018vizwiz}.  Accordingly, we now examine the benefit of our large-scale quality dataset for training algorithms to detect when images are unrecognizable and so not captionable.  

\subsection{Motivation: Inadequate Existing Methods}
Before exploring novel algorithms, it is important to first check whether existing methods are suitable for our purposes.  Accordingly, we check whether related NR-IQA systems can detect when images are unrecognizable.  To do so, we apply five NR-IQA methods on the complete VizWiz-QualityIssues dataset: BRISQUE \cite{mittal2012no}, NIQE \cite{mittal2012making}, CNN-NRIQA \cite{kang2014convolutional}, DNN-NRIQA \cite{bosse2017deep}, and NIMA \cite{talebi2018nima}. The first two are popular conventional methods that rely on hand-crafted features.  The last three are based on neural networks and trained on IQA datasets mentioned in Section \ref{sec:related_work}. For example, DNN-NRIQA-TID and DNN-NRIQA-LIVE in Figure~\ref{fig:NRIQA} are trained on the TID dataset and LIVE dataset, respectively.  Intuitively, if the algorithms are effective for this task, we would expect that the scores for recognizable images are distributed mostly in the high-score region, while the scores for unrecognizable images are distributed mostly in the low-score region.  

Results are shown in Figure~\ref{fig:NRIQA}.  A key finding is that the distributions of scores for recognizable and unrecognizable images heavily overlap.  That is, none of the methods can distinguish recognizable images from unrecognizable images in our dataset.  This finding shows that existing methods trained on existing datasets (i.e., LIVE, TID, CSIQ) are unsuitable for our novel task on the VizWiz-QualityIssues dataset.  This is possibly in part because quality issues resulting from artificial distortions, such as compression, Gaussian blur, and additive Gaussian noise, differ from natural distortions triggered by poor camera focus, lighting, framing, etc.  This also may be because there is no 1-1 mapping between scores indicating overall image quality and our proposed task, since an image with a low quality score may still have recognizable content. 

\subsection{Proposed Algorithm} 
\label{sec:recognizability_exp}
Having observed that existing IQA methods are inadequate for our problem, we now introduce models for our novel task of assessing whether an image is recognizable.  

\vspace{-0.7em}
\paragraph{Architecture.} We use ResNet-152 \cite{he2016deep} to extract image features, which are then processed by 2-dimensional global-pooling followed by two fully connected layers. The final layer is a single neuron with a sigmoid activation function.\footnote{Due to space constraints, we demonstrate the effectiveness of this architecture for assessing the quality flaws in the Supplementary Materials.  The primary difference for that architecture is that we replace ResNet-152 with XceptionNet \cite{chollet2017xception}, use three fully connected layers, and a final layer of eight neurons with eight sigmoid functions.}  We train this algorithm using an Adam optimizer with the learning rate set to 0.001 for 8 epochs.  We fix the ResNet weights pre-trained on ImageNet \cite{deng2009imagenet} and only learn the weights in the two fully connected layers.    

\vspace{-0.7em}
\paragraph{Dataset Splits.}
For training and evaluation of our algorithm, we apply a 52.5\%/37.5\%/10\% split to our dataset to create the training, validation, and test splits.  

\begin{table}[b!]
    \centering
    \begin{tabular}{c c c c}
        \toprule
             & Avg. precision & Recall & F1 \\
        \midrule
            ResNet-152 &  80.0 & \textbf{75.1} & \textbf{71.2}\\
            Random guessing & 16.6 & 14.6 & 15.5\\
            SIFT & \textbf{87.2} & 42.3 & 56.9 \\
            HOG + linear SVM & 56.4 & 41.2 & 47.6 \\
        \bottomrule
    \end{tabular}
    \caption{Performance of algorithms in assessing whether image content can be recognized (and so captioned).}
    \label{tab:recognizability}
\end{table}

\begin{table*}[t!]
  \centering
        \begin{tabular}{ c  c  c  c  c  c  c  c  c  c}
    
    \toprule
       && \bf B@1 & \bf B@2 & \bf B@3 & \bf B@4 & \bf METEOR & \bf ROUGE-L & \bf CIDEr-D & \bf SPICE  \\
     \midrule
    \multirow{4}{*}{{{\bf AoANet ~\cite{huang2019attention}}}} 
        & full training set & 63.3 & 44.3 & 29.9 & 19.7 & 18.0 & 44.4 & 43.6 & 11.2  \\ 
        & perfect flag & 63.3 & 43.8 & 29.5 & 19.9 & 18.1 & 44.2 & 43.6 & 11.5 \\
        & predicted flag & 63.2 & 44.0 & 29.5 & 19.8 & 18.1 & 44.2 & 42.9 & 11.5 \\
        & random sample & 62.5 & 43.3 & 28.8 & 18.9 & 18.0 & 44.1 & 41.9 & 11.4 \\
    \hline
    \multirow{4}{*}{{{\bf SGAE~\cite{yang2019auto}}}} 
       & full training set & 62.8 & 43.3 & 28.6 & 18.8 & 17.3 & 44.0 & 32.4 & 10.4 \\ 
        & perfect flag & 63.0 & 43.1 & 28.6 & 18.9 & 17.2 & 43.9 & 32.5 & 10.3 \\
        & predicted flag & 63.1 & 43.1 & 28.4 & 18.7 & 17.2 & 44.0 & 32.4 & 10.4 \\
        & random sample & 62.4 & 42.7 & 27.9 & 18.2 & 17.1 & 43.7 & 30.4 & 10.4 \\
    \bottomrule
    \end{tabular}
          \vspace{0.1em}
        \caption{Performance of two image captioning algorithms with respect to eight metrics trained on the full captioning-training-set, training images annotated to be recognizable (perfect flag), training images predicted to be recognizable (predicted flag), and a subset random sampled from the captioning-training-set. (B@ = BLEU-)}
        ~\label{tab:captioning_scratch}
        \vspace{-0.7em}
\end{table*} 

\vspace{-0.7em}
\paragraph{Baselines.}
We compare our algorithm to numerous baselines.  Included is random guessing, which means an image is unrecognizable with probability $0.148$.  We also analyze a linear SVM that predicts with scale-invariant feature transform (SIFT) features.  Intuitively, a low-quality image should have few/no key points.  We also evaluate a linear SVM that predicts from histogram of oriented gradients (HOG) features.  

\vspace{-0.7em}
\paragraph{Evaluation Metrics.}
We evaluate each method using average precision, recall, and f1 scores.  Accuracy is excluded because the distributions of unrecognizability are highly biased to ``false" and such unbalanced data suffer from the accuracy paradox.

\vspace{-0.7em}
\paragraph{Results.}
Results are shown in Table~\ref{tab:recognizability}. We observe that both SIFT and HOG are much stronger baselines than random guessing and get high scores on precision, especially $87.2$ for SIFT. However, they both get low scores on recall. This means that SIFT and HOG are good at capturing a subset of unrecognizable images but still miss many others. On the other hand, the ResNet model gets much higher recall scores while maintaining decent average precision scores, implying that it is more effective at learning the characteristics of unrecognizable images.\footnote{Again, due to space constraints, results showing prediction performance for quality flaw classification is in the Supplementary Materials.}  This is exciting since such an algorithm can be of immediate use to blind photographers, who otherwise must wait nearly two minutes to learn their image is unsuitable quality for image captioning. 

\subsection{Application: Efficient Dataset Creation} 
\label{sec_budgetAllocation}
We now examine another potential benefit of our algorithm in helping to create a large scale training dataset.  

To support this effort, we divide the dataset into three sets.  One set is used to train our image unrecognizability algorithm.  A second set is used to train our image captioning algorithms, which we call the captioning-training-set.  The third set is used to evaluate our image captioning algorithms, which we call the captioning-evaluation-set.  

We use our method to identify which images in the captioning-training-set to use for training image captioning algorithms.  In particular, the $N$ images flagged as recognizable are included and the remaining images are excluded.  We compare this method to three baselines, specifically training on: all images in the captioning-training-set, a random sample of $N$ images in the captioning-training-set, a perfect sample of $N$ images in the captioning-training-set that are known to be recognizable images.

We evaluate two state-of-art image captioning algorithms, trained independently on each training set, with respect to eight evaluation metrics:  BLEU-1-4~\cite{papineni2002BLEUmethodautomatic}, METEOR~\cite{denkowski:lavie:meteor-wmt:2014}, ROUGE-L~\cite{lin2004rouge}, CIDEr-D~\cite{vedantam2015CiderConsensusbasedimage}, and SPICE~\cite{anderson2016SpiceSemanticpropositional}. 

Results are shown in Table \ref{tab:captioning_scratch}.  Our method performs comparably to when the algorithms were trained on all images as well as the perfect set.  In contrast, our method yields improved results over the random sample.  Altogether, these findings offer promising evidence that our prediction system is successfully retaining meaningful images while removing images that are not informative for the captioning task (i.e., unrecognizable).  This reveals that a benefit of using the recognizability prediction system is to save time and money when crowdsourcing captions (by first removing unrecognizable images), without diminishing the performance of downstream trained image captioning algorithms.
\section{Recognizing Unanswerable Visual Questions}
\label{sec_vqa}
\begin{figure}[t!]
    \centering
    \includegraphics[width=\linewidth]{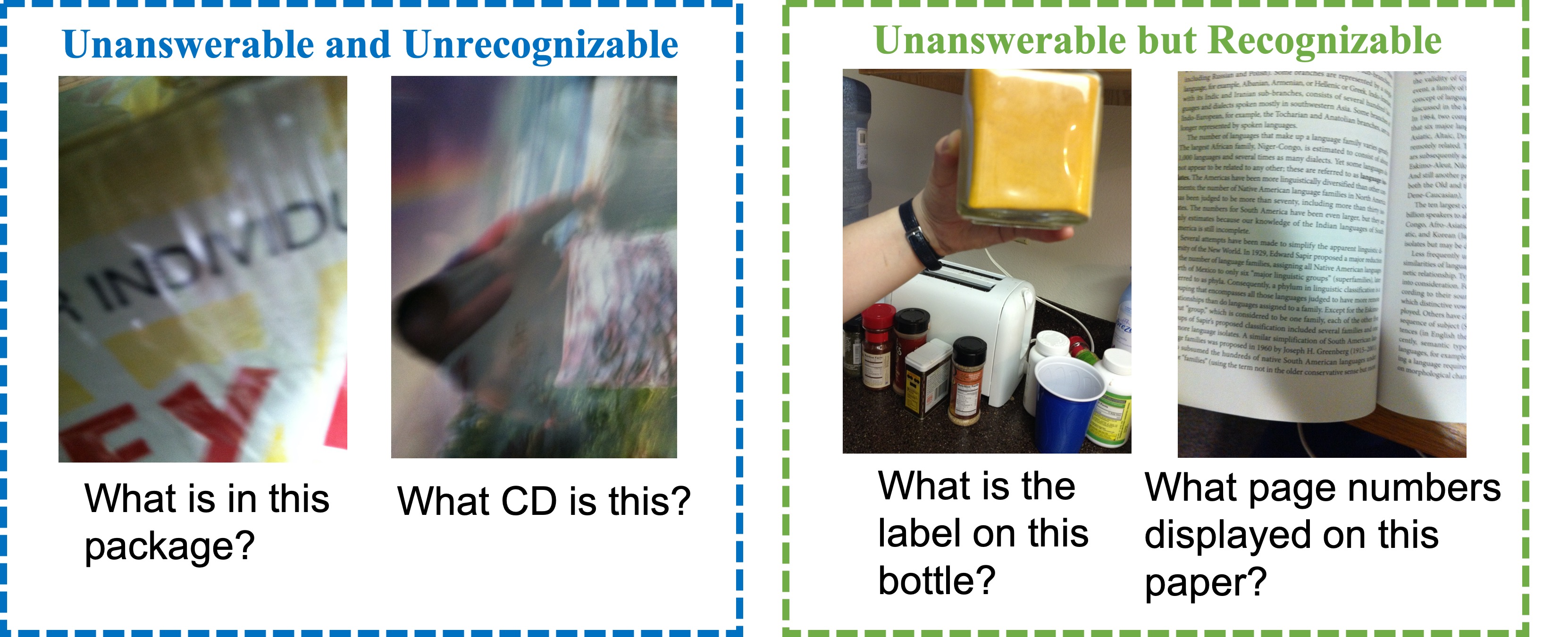}
    \caption{Examples of visual questions that are unanswerable for two reasons. The left two examples have unrecognizable images while the right two examples have recognizable images but the content of interest is missing from the field of view. Our posed algorithm correctly predicts why visual questions are unanswerable for these examples.}
    \label{fig:vqa}
\end{figure}

The visual question ``answerability" problem is to decide whether a visual question can be answered~\cite{gurari2018vizwiz}.  Yet, as exemplified in Figure~\ref{fig:vqa}, visual questions can be unanswerable because the image is unrecognizable or because the answer to the question is missing in a recognizable image.
Towards enabling more fine-grained guidance to photographers regarding how to modify the visual question so it is answerable, we move beyond predicting \emph{whether} a visual question is unanswerable~\cite{gurari2018vizwiz} and introduce a novel problem of predicting \emph{why} a visual question is unanswerable.

\subsection{Motivation}
\begin{figure}
     \centering
     \begin{subfigure}[b]{0.48\textwidth}
         \centering
         \includegraphics[width=\textwidth]{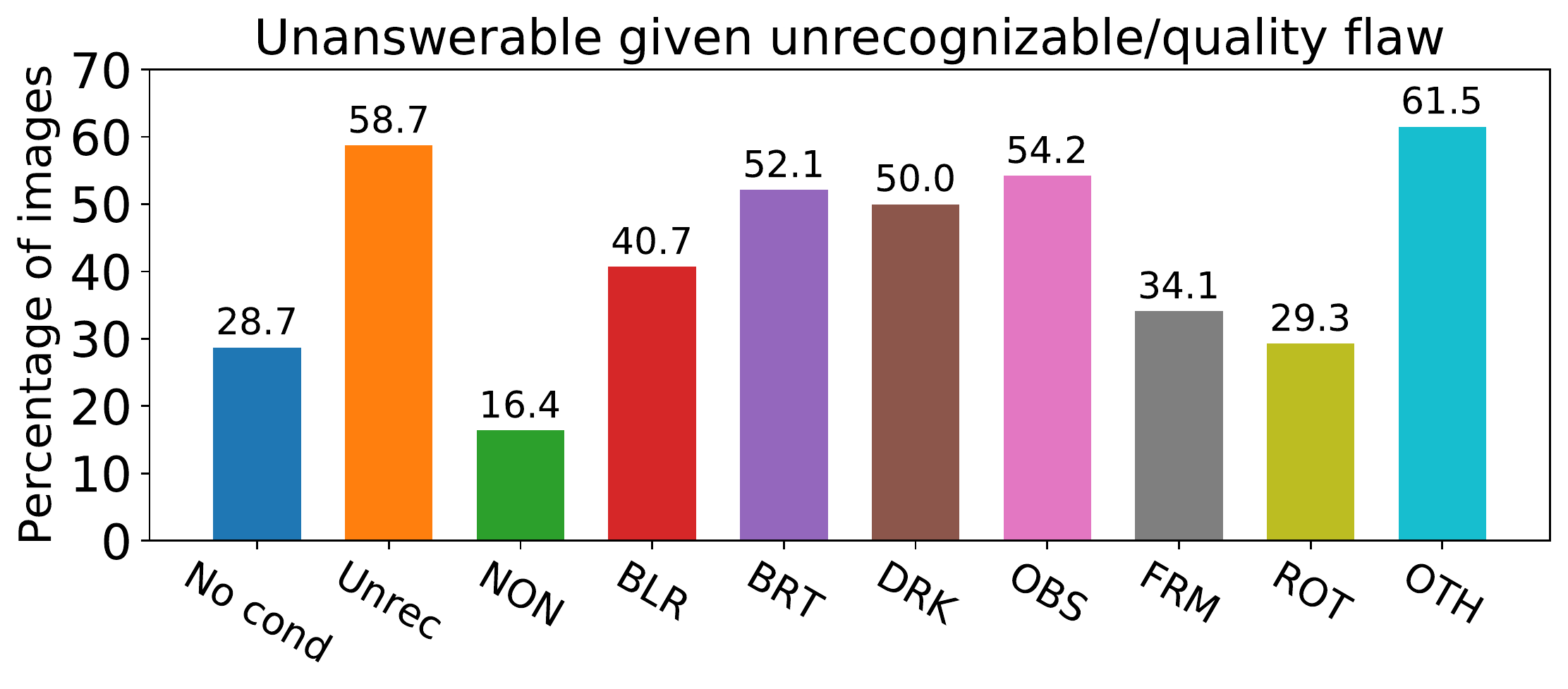}
     \end{subfigure}
     \\
     \begin{subfigure}[b]{0.48\textwidth}
         \centering
         \includegraphics[width=\textwidth]{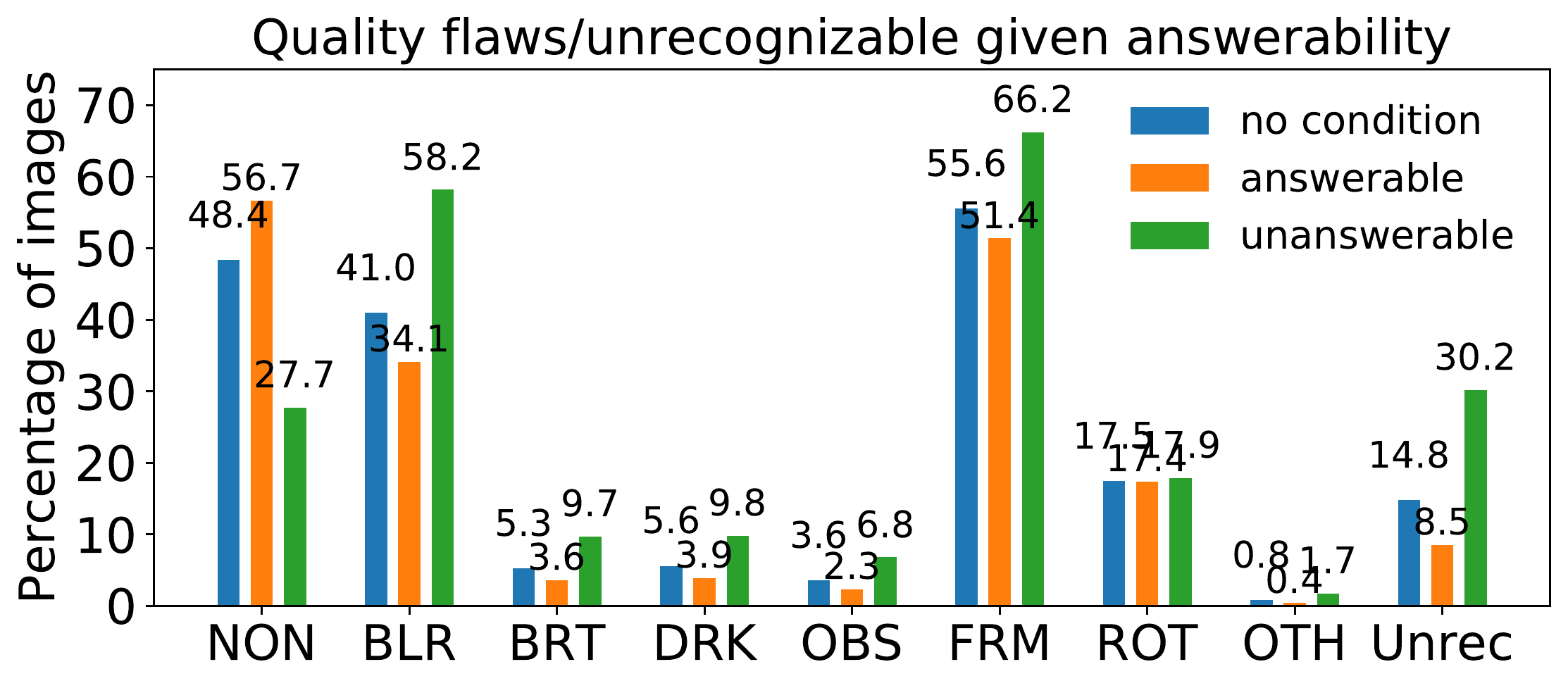}
     \end{subfigure}\vspace{-1ex}
     \caption{Top: Fractions of unanswerable questions conditioned on unrecognizability or a quality flaw. Bottom: Fractions of quality issues and unrecognizable images given answerability. Values are scaled by being multiplied with 100.}
    \label{fig:unans_and_quality}
\end{figure}

We extend the VizWiz-VQA dataset~\cite{gurari2018vizwiz}, which labels each image-question pair as answerable or unanswerable.  We inspect how answerability relates to recognizability and each quality flaw. For convenience, we use the following notations: $A$: answerable, $\bar{A}$: unanswerable, $R$: recognizable: $\bar{R}$: unrecognizable, $Q$: quality issues, and $P(\cdot)$: probability function.  Results are shown in Figure~\ref{fig:unans_and_quality}. We can observe that for most quality flaws $Q$, $P(\bar{A}|Q)$ is larger than $P(\bar{A})$, and $P(\bar{A})=28.7\%$ increases to $P(\bar{A}|\bar{R})=58.7\%$. Additionally, the probability $P(\bar{R})$ increases from $14.8\%$ to $P(\bar{R}|\bar{A})=30.2\%$ when questions are known to be unanswerable. Observing that a large reason for unanswerable questions is that images are unrecognizable images, we are motivated to equip VQA systems with a function that is able to clarify why their questions are unanswerable.

\subsection{Proposed Algorithm}
\paragraph{Algorithm.}
Our algorithm extends the Up-Down VQA model \cite{anderson2018bottom}.  It takes as input encoded image features and a paired question.  Image features could be grid-level features extracted by ResNet-152 \cite{he2016deep} as well as object-level features extracted by Faster-RCNN \cite{ren2015faster} or Detectron \cite{Detectron2018, wu2019detectron2}. The input question is first encoded by a GRU cell. Then, a top-down attention module computes a weighted image feature from the encoded question representation and the input image features. The image and question features are coupled by element-wise multiplication. This coupled feature is processed by the prediction module to predict answerability and recognizability.  We employ two different activation functions at the end of the model to make the final prediction. The first one is softmax which predicts three exclusive classes: answerable, unrecognizable, and insufficient content information (answers cannot be found in images).  The second activation function is two independent sigmoids, one for answerability and the other for recognizability.  We train the network using an Adam optimizer with a learning rate of 0.001, only for the layers after feature extraction.

\vspace{-0.7em}
\paragraph{Dataset Splits.}
We split VizWiz dataset into training/validation/test sets according to a 70\%/20\%/10\% ratio.  

\vspace{-0.7em}
\paragraph{Evaluation Metrics.}
We evaluate performance using average precision, precision, recall, and f1 scores, for which a simple threshold $0.5$ is used to binarize probability values. For inter-model comparisons, we also report the precision-recall curve for each variant. 

\vspace{-0.7em}
\paragraph{Baselines.}
For comparison, we consider a number of baselines.  One approach is the original model for predicting whether a visual question is answerable, and also employs a top-down attention model~\cite{gurari2018vizwiz}. We also evaluate the random guessing, SIFT, and HOG baselines used to evaluate the recognizability algorithms in the previous section.

\begin{table}[t!]
    \centering
    \resizebox{\columnwidth}{!}{
    \begin{tabular}{l *6c}
    \toprule
         {} & \multicolumn{3}{c}{Unans} & \multicolumn{3}{c}{Unrec given unans} \\
    \cmidrule(lr){2-4}
    \cmidrule(lr){5-7}
        {} & AP & Rec & F1 & AP & Rec & F1\\
    \midrule
        \cite{gurari2018vizwiz} & 71.7 & $\bm{-}$ & 64.8 & $\bm{-}$ & $\bm{-}$ & $\bm{-}$ \\
    \hdashline
        Rand guess & $\bm{-}$ & $\bm{-}$ & $\bm{-}$ & 31.1$^\ast$ & 14.8 & 20.0 \\
        SIFT & $\bm{-}$ & $\bm{-}$ & $\bm{-}$ & \textbf{94.9}$^\ast$ & 45.3 & 61.3 \\
        HOG & $\bm{-}$ & $\bm{-}$ & $\bm{-}$ & 73.1$^\ast$ & 44.9 & 55.7 \\
    \hdashline
        TD+soft & 72.6 & 77.3 & 67.0 & 82.2 & \textbf{79.3} & 75.0 \\
        TD+sigm & 73.6 & 71.2 & \textbf{68.0} & 86.6 & \textbf{79.3} & 78.6 \\
        BU+sigm & 73.0 & 66.6 & 66.7 & 87.4 & 73.7 & 78.7 \\
        TD+BU+sigm& \textbf{74.0} & \textbf{82.3} & 67.9 & 87.7 & \textbf{79.3} & \textbf{79.7} \\
        sigm w/o att. & 67.7 & 66.1 & 64.2 & 86.7 & 66.7 & 74.2 \\
    \bottomrule
    \end{tabular}
    }
    \begin{tablenotes}
        \item \textbf{TD}: top-down attention. \textbf{BU}: bottom-up attention. \textbf{soft}: softmax. \textbf{sigm}: sigmoid. \textbf{att}: attention. \textbf{AP}: average precision. \textbf{Rec}: recall. \textbf{Unrec}: Unrecognizable. \textbf{Unans}: Unanswerable.  
        \item $^\ast$: Precision is calculated, since true or false is predicted instead of a probability. 
    \end{tablenotes}
    \caption{Performance of predicting why a visual question is unanswerable: unrecognizable image versus unanswerable because the content of interest is missing from the field of view.  \cite{gurari2018vizwiz} only predicts answerability and serves as the baseline for unanswerability prediction. Random guessing, SIFT, and HOG only predict recognizability and serve as the baselines for unrecognizability prediction.}
    \label{tab:benchmarking}
\end{table}

\begin{figure}[t!]
    \centering
    \includegraphics[width=0.8\linewidth]{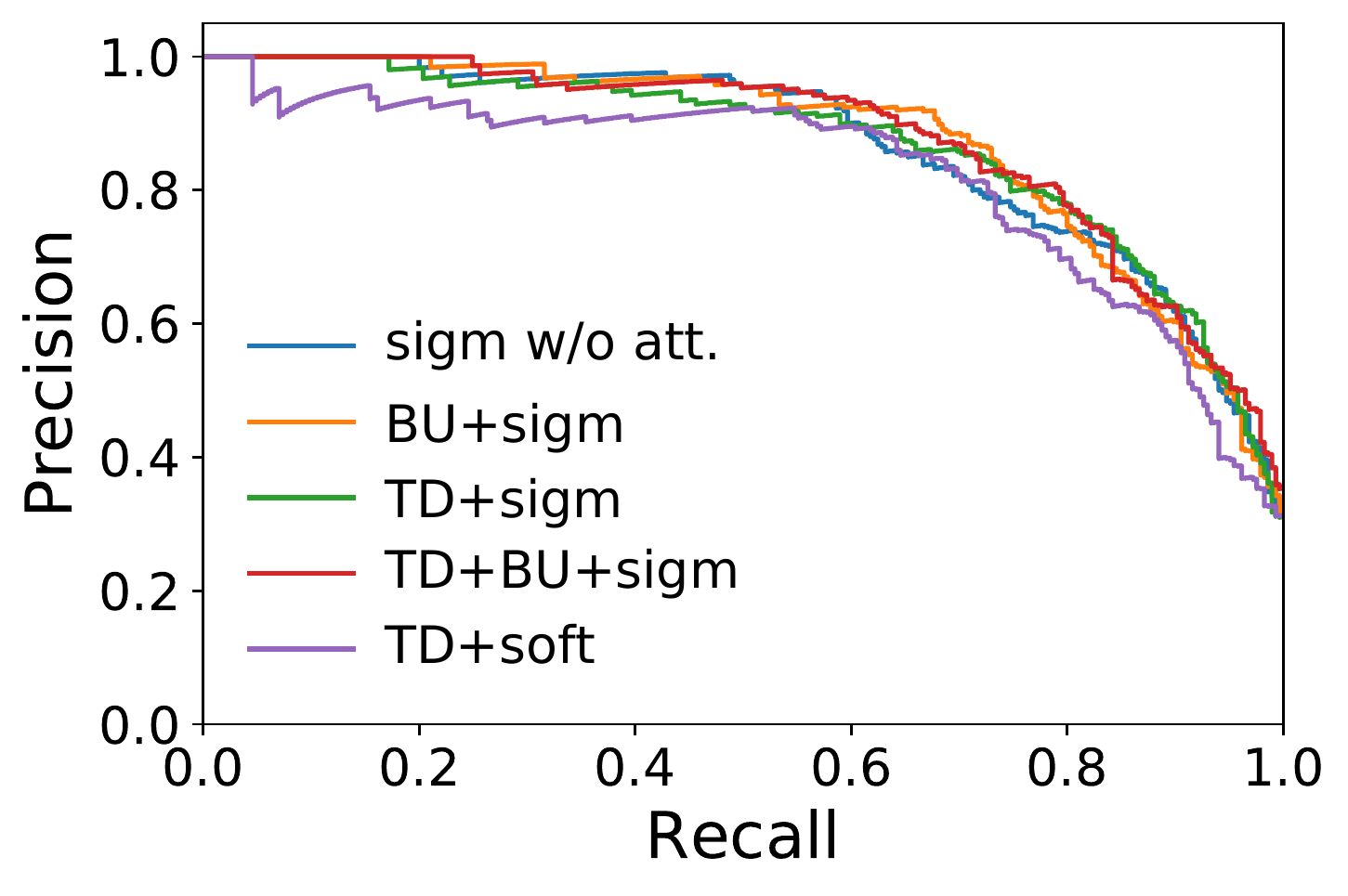}\vspace{-2ex}
    \caption{Precision-recall curves for five algorithms predicting unrecognizability when questions are unanswerable.}
    \label{fig:precision_recall}
\end{figure}

\paragraph{Results.}
Results are shown in Table~\ref{tab:benchmarking} and Figure~\ref{fig:precision_recall}.  Our models perform comparably to the answerability baseline \cite{gurari2018vizwiz}.  This is exciting because it shows that jointly learning to predict answerability with recognizability does not degrade the performance; i.e., the average precision scores from TD+softmax and TD+sigmoid models are better than the one from the baseline~\cite{gurari2018vizwiz} ($72.6$, $73.6 > 71.7$) as well as the F1 scores ($67.0$, $68.0 > 64.8$).

Our results also highlight the importance of learning to predict jointly the answerability with recognizability task (i.e., rows 5--9) over relying on more basic baselines (i.e., rows 2--4).  As shown in Table~\ref{tab:benchmarking}, low recall values imply that SIFT and HOG fail to capture many unrecognizable images, while our models learn image features and excel in recall and f1 scores.

Next, we compare the results from TD+softmax and TD+sigmoid.  We observe they are comparable in unanswerability prediction due to comparable average precision scores and F1 scores. For unrecognizability prediction, TD+softmax is a bit weaker than TD+sigmoid because due to slightly lower average precision and F1 scores. One reason for this may be the manual assignment of unrecognizability to false when answerability is true. Originally, $14.8\%$ of images are unrecognizable, but after assignment, the portion drops to $8.7\%$. Learning from more highly biased data is a harder task, which could in part explain the weaker performance of TD+softmax model.
\section{Conclusions}
We introduce a new image quality assessment dataset that emerges from an authentic use case where people who are blind struggle to capture high-quality images towards learning about their visual surroundings. We demonstrate the potential of this dataset to encourage the development of new algorithms that can support real users trying to obtain image captions and answers to their visual questions.  The dataset and all code are publicly available at \texttt{https://vizwiz.org}.

\noindent
\paragraph{Acknowledgements.}
We gratefully acknowledge funding support from the National Science Foundation (IIS-1755593), Microsoft, and Amazon.  We thank Nilavra Bhattacharya and the crowdworkers for their valuable contributions to creating the new dataset.

\balance{}
{\small
\bibliographystyle{ieee}
\bibliography{myReferences}
}

\section*{Appendix}
This document supplements Sections 3 and 4 of the main paper.  In particular, it includes the following:

\begin{itemize}
\item Details and motivation for quality flaw interrelation index (supplements \textbf{Section 3.2}). 
\item Result of quality flaw prediction (supplements \textbf{Section 4.2}).
\item Figures illustrating the crowdsourcing interface used to curate our labels (supplements \textbf{Section 3.1}), diversity of resulting unrecognizable images (supplements \textbf{Section 3.2}), performance of our prediction system in classifying unrecognizable images (supplements \textbf{Section 4.2}), and performance of the prediction of the reason for unanswerable questions (supplements \textbf{Section 5.2}).
\item Clarification about baselines used for \textbf{Section 4.3}.
\end{itemize}

\section{Quality flaw interrelation index}
\paragraph{Details and motivation}
The most straightforward way to explore the relation of two quality flaws $A$ and $B$ is to look at their the co-occurrence or their joint probability $P(A,B)$. However, $P(A,B)$ cannot really capture the interrelation between quality flaws. For instance, we cannot say that the relation between DRK and FRM is stronger than the one between DRK and OBS simply because of $P(\text{DRK}, \text{FRM}) \gg P(\text{DRK}, \text{OBS})$. The reason for $P(\text{DRK}, \text{FRM}) \gg P(\text{DRK}, \text{OBS})$ is actually due to $P(\text{FRM})=55.6\% \gg P(\text{OBS})=3.6\%$ but has nothing to do with the interrelation of quality flaws. 

Consequently, we introduce a new measure which we call interrelation index $I(A,B)$, which is defined as follows:
\begin{equation}
    I(A,B)=\frac{P(B|A)}{P(B)} - \frac{P(B|\bar{A})}{P(B)}.
    \label{eq:interrelation_supp}
\end{equation}
There are several advantages of this measure:
\begin{enumerate}
    \item It measures causality from $A$ to $B$: we can show that if $P(A)$ and $P(B)$ are both greater than zero, either $P(B|A) \geq P(B) \geq P(B|\bar{A})$ or $P(B|A) < P(B) < P(B|\bar{A})$ holds. Therefore, if $I(A,B) > 0$, then the existence of A must trigger B to happen more (i.e., $P(B|A) \geq P(B)$) and the inexistence of A must make B happen less (i.e., $P(B) \geq P(B|\bar{A})$), and vice versa. 
    \item It measures co-occurence of $A$ and $B$: We can show that if $P(B|A) \geq P(B) \geq P(B|\bar{A})$, then $P(A|B) \geq P(A) \geq P(A|\bar{B})$ (it is also true for $<$ sign). Hence, we have $I(A,B) > 0 \Leftrightarrow I(B,A) > 0$. In other words, if $A$ makes $B$ happen more often, then $B$ must make $A$ happen more as well, and vice versa. 
    \item It avoids the aforementioned problem of using joint probability. That is, if $P(A) \gg P(B)$, it is very likely $P(A,C) \gg P(B,C)$. However, which of the values of $I(A,C)$ and $I(B,C)$ is greater and how greater it is cannot be told from $P(A) \gg P(B)$.
\end{enumerate}

\begin{figure*}[!t]
    \centering
    \includegraphics[width=0.93\textwidth]{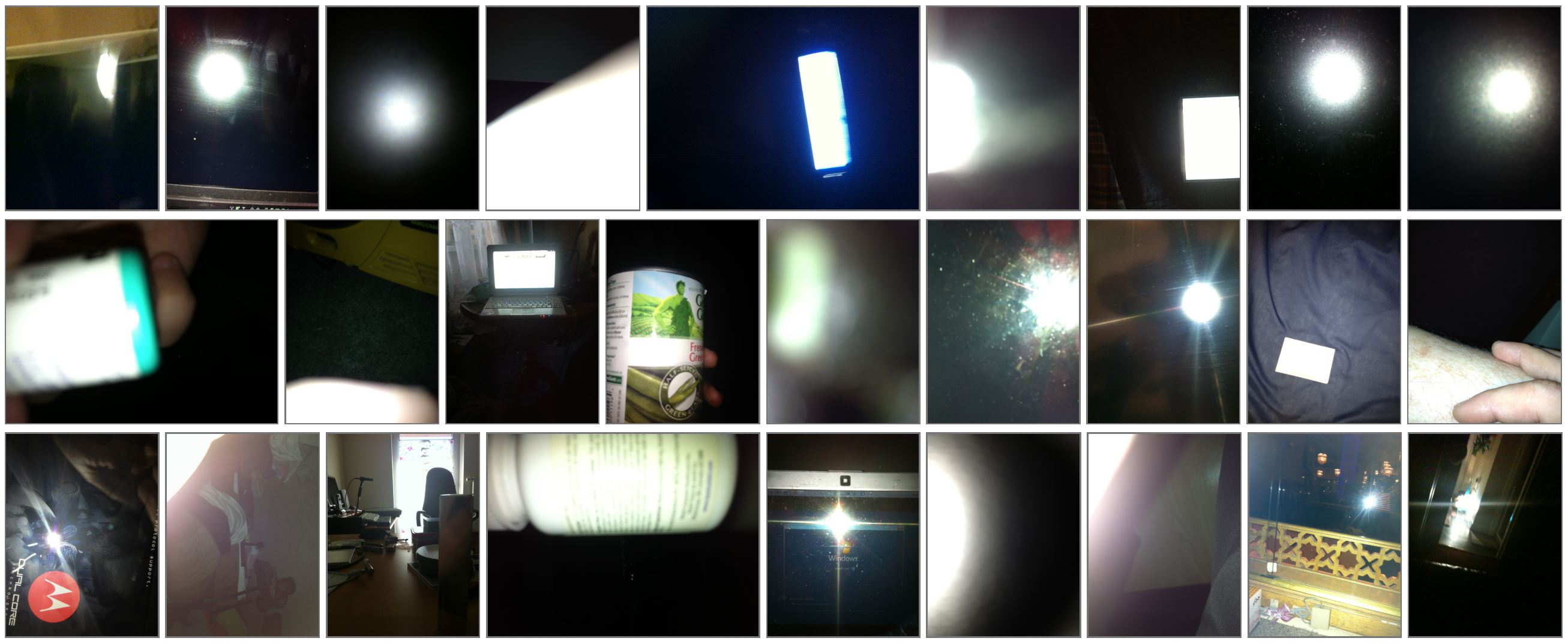}
    \caption{Examples of images that are both too dark and too bright. Note that both recognizable and unrecognizable images can appear here, since quality flaws do not necessarily render an image unrecognizable.}
    \label{fig:drk_brt}
\end{figure*}

\paragraph{Co-occurrence of DRK and BRT.} Since the values $I(\text{DRK},\text{BRT}) = 74$ and $I(\text{BRT},\text{DRK}) = 73$ are both greater than zero, it means that the quality flaws of DRK and BRT tend to co-occur despite their contradictory concepts. Nevertheless, the examples of such images in Figure~\ref{fig:drk_brt} explain why this phenomenon happens. The main reason for this phenomenon is when blind people take pictures in places with poor lighting, they are not aware that the flashlights on mobile devices are turned on automatically, and therefore pictures taken are usually of dark surroundings and a bright spot. Note that this phenomenon is not captured by the joint probability of DRK and BRT, since $P(\text{DRK},\text{BRT}) = 0.53\%$ is an extremely small value which does not manifest too much. 

\begin{figure*}[!t]
    \centering
    \includegraphics[width=0.93\textwidth]{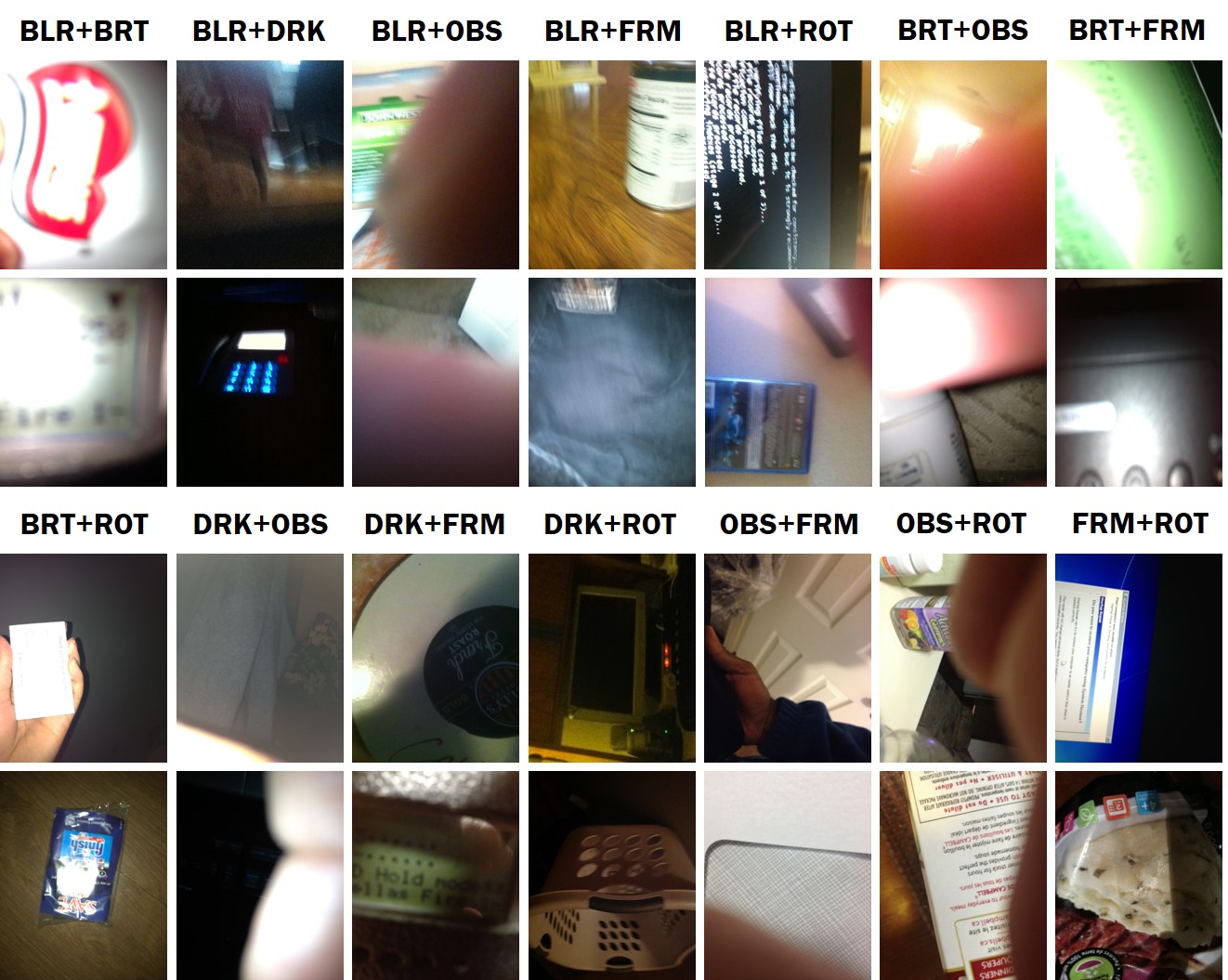}
    \caption{Examples of the co-occurrence of all quality flaw pairs.  Again we obsere both recognizable and unrecognizable images appear since quality flaws do not necessarily render an image unrecognizable.}
    \label{fig:quality_flaw_pair}
\end{figure*}

\paragraph{Co-occurrence of quality flaws.} We exemplify the co-occurrence of other pairs of quality flaws in Figure~\ref{fig:quality_flaw_pair}.

\begin{table*}[h!]
  \centering
        \begin{tabular}{ c  c  c  c  c  c  c  c  c  c}
    
    \toprule
       && \bf NON & \bf BLR & \bf BRT & \bf DRK & \bf OBS & \bf FRM & \bf ROT & \bf OTH  \\
     \midrule
    \multirow{3}{*}{{\bf Xception }} 
        & precision & 72.9 & 80.1 & 62.9 & 58.5 & 53.6 & 77.0 & 72.6 & 60.0  \\ 
        & recall & 79.0 & 80.1 & 49.8 & 57.3 & 39.7 & 82.4 & 69.8 & 9.1 \\
        & f1 score & 75.8 & 80.1 & 55.6 & 57.9 & 45.6 & 79.6 & 71.2 & 15.8 \\
    \hline
    \multirow{3}{*}{{\bf Random guessing}} 
       & precision & 48.6 & 40.5 & 4.9 & 7.2 & 4.0 & 55.0 & 15.6 & 0.0 \\ 
        & recall & 50.5 & 40.3 & 4.3 & 6.7 & 4.0 & 54.3 & 15.7 & 0.0 \\
        & f1 score & 49.5 & 40.4 & 4.5 & 7.0 & 4.0 & 54.6 & 15.6 & 0.0 \\
    \bottomrule
    \end{tabular}
          \vspace{-0.5em}
        \caption{Performance of quality flaw prediction}
        ~\label{tab:quality_flaw_prediction}
        \vspace{-1.5em}
\end{table*}

\begin{figure}
    \centering
    \includegraphics[width=1.0\linewidth]{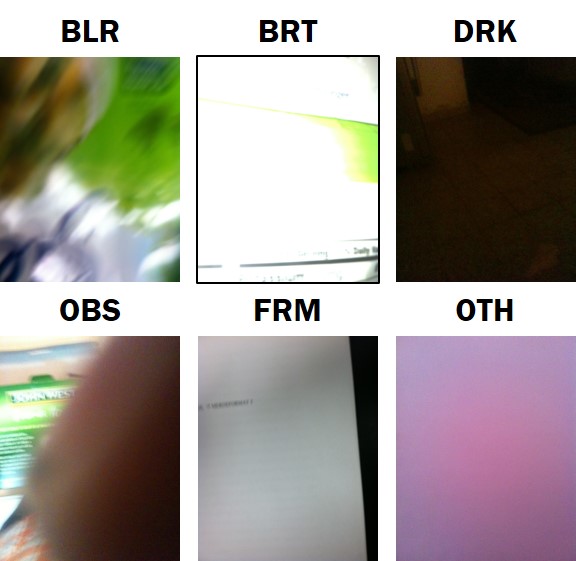}
    \caption{Unrecognizable images due to different quality flaws.}
    \label{fig:uncaptionable_flaw}
\end{figure}

\section{Quality flaw prediction}
Performance of quality flaw classification is shown in Table~\ref{tab:quality_flaw_prediction}. We can tell that the Xception model outperforms the random guessing baseline for each quality flaw, with respect to precision, recall, and f1 score. Furthermore, Xception predicts much better in NON, BLR and FRM flaws for large portions of the dataset. On the other hand, quality flaws that represent small portions of the dataset are prone to few-shot learning, and so learning to predict them is harder. In the extreme case of OTH, with it representing $0.8\%$ of the data, the Xception model yields very poor scores of $9.1$ and $15.8$ for recall and f1 score, respectively.

\section{Miscellaneous}
\begin{figure*}[t]
\centering
\includegraphics[width=0.9\textwidth]{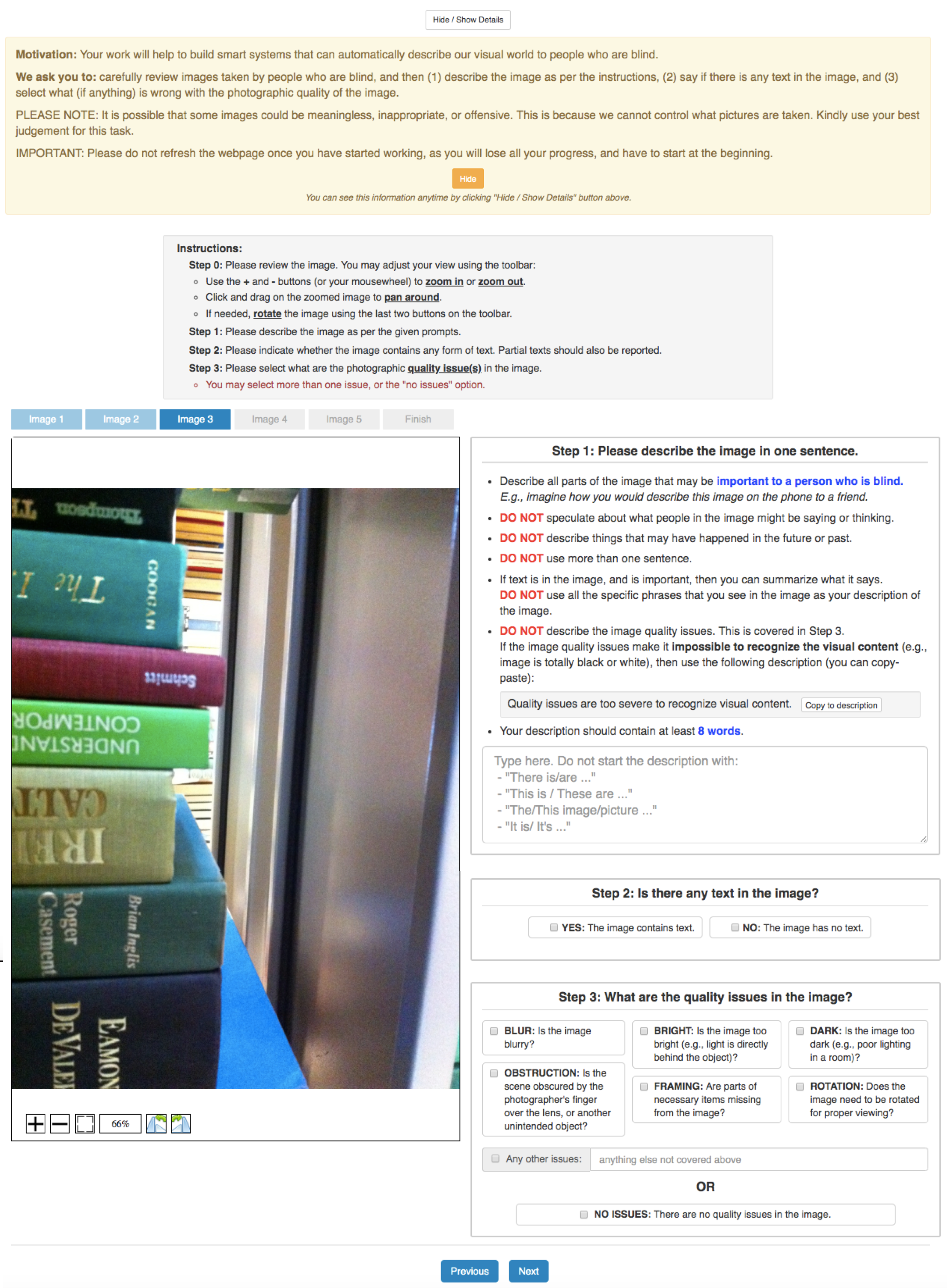}
\caption{Interface used to crowdsource the collection of image captions.}
\label{fig:taskInterface}
\end{figure*}

\begin{figure*}[t!]
\centering
\includegraphics[width=1.0\textwidth]{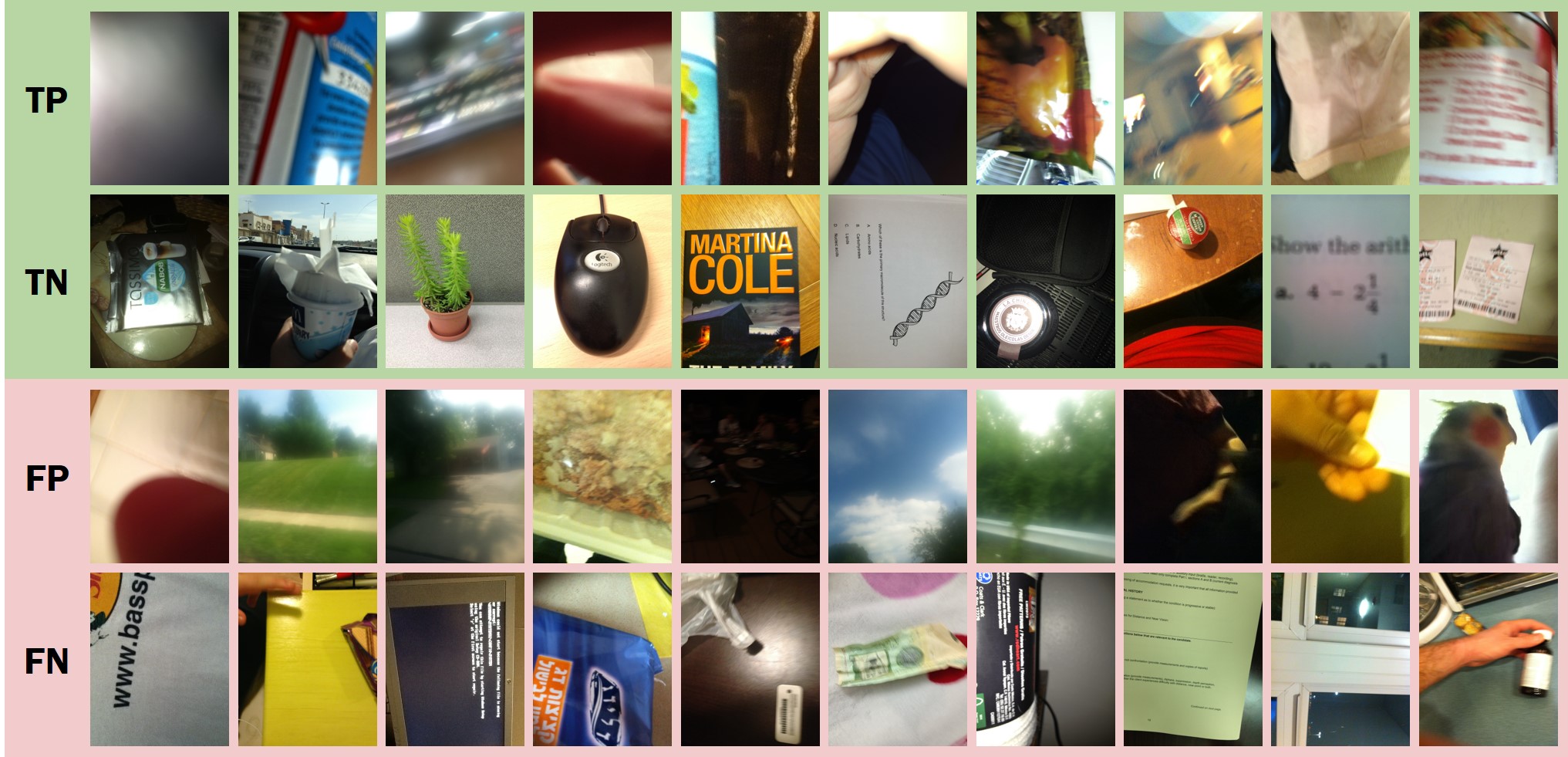}
\vspace{-1.5em}
\caption{Examples of true-positives (TP), true-negatives (TN), false-positives (FP), and false-negatives (FN) in unrecognizability prediction. \textbf{TP}: unrecognizable images predicted to be unrecognizable. \textbf{TN}: recognizable images predicted to be recognizable. \textbf{FP}: recognizable images predicted to be unrecognizable. \textbf{FN}: unrecognizable images predicted to be recognizable.}
\vspace{-1.5em}
\label{fig:uncaptionability_pred}
\end{figure*}

\begin{figure*}[!t]
\centering
\includegraphics[width=1.0\textwidth]{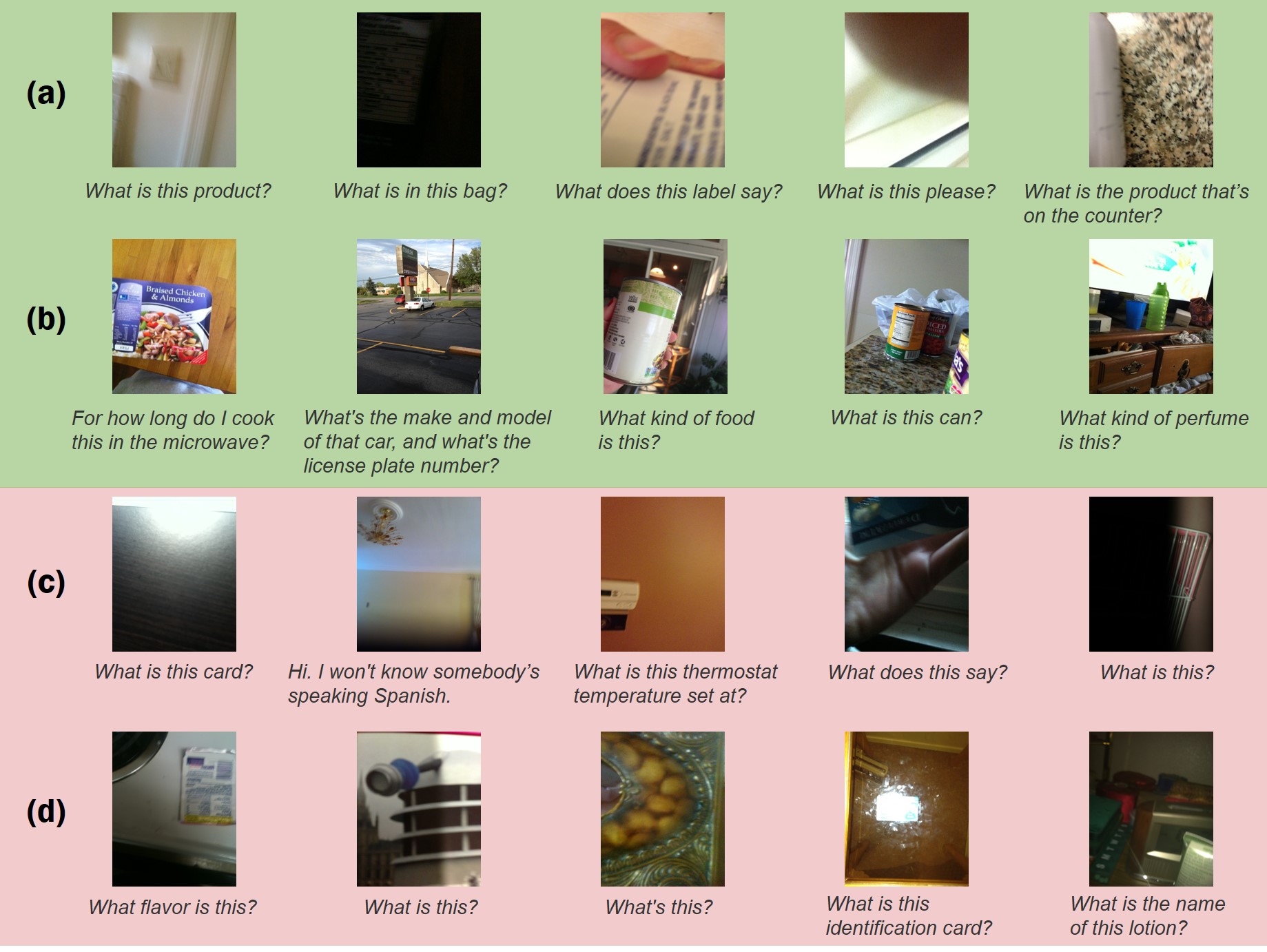}
\vspace{-1.5em}
\caption{Prediction of the reason for unanswerable questions. Note that each visual question pair here is unanswerable. (a) Unanswerable questions are due to unrecognizable images and so are predictions. (b) Unanswerable questions are due to insufficient content and so are predictions. (c) Unanswerable questions are due to insufficient content but predicted to be due to unrecognizable images. (d) Unanswerable questions are due to unrecognizable images but predicted to be due to insufficient content.}
\vspace{-1.5em}
\label{fig:uncap_pred_given_unans}
\end{figure*}

\begin{itemize}
    \item Figure~\ref{fig:uncaptionable_flaw} illustrates the diversity of unrecognizable images that can arise from different quality flaws.
    \item Figure~\ref{fig:taskInterface} shows a screen shot of the crowdsourcing interface used to collect the labels for the dataset.  
    \item Figure~\ref{fig:uncaptionability_pred} shows the examples of unrecognizability prediction by the Xception model.
    \item Figure~\ref{fig:uncap_pred_given_unans} shows the examples of the prediction of the reason for unanswerable questions. The prediction model used is ``TD+sigmoid" model.
\end{itemize}

\section{Section 4.3: Clarification about Baselines}
The two baselines, ``random flag" and ``perfect flag", use the same number of images from the captioning-training-set as our method for algorithm training.  That count is determined by our predictor, specifically the number of images that remain after removing all images that are deemed to be unrecognizable.  ``Random flag" chooses a random sample from the captioning-training-set.  ``Perfect flag" chooses images based on a ranking of images based on how many crowdworkers flag the images as unrecognizable, with selection starting from those where all five crowdworkers agreed the image is unrecognizable.

\end{document}